# Distributed adaptive algorithm based on the asymmetric cost of error functions

Sihai Guan[1,2], Qing Cheng[3], Yong Zhao[4]

(1. College of Electronic and Information, Southwest Minzu University, Chengdu, China. 2. Key Laboratory of Electronic and Information Engineering, State Ethnic Affairs Commission, Chengdu, China. 3. Sichuan Vocational College of Finance and Economics, Chengdu, China. 4. School of Mechanical and Power Engineering, Henan Polytechnic University, Jiaozuo, China.)

**Abstract**: In this paper, a family of novel diffusion adaptive estimation algorithm is proposed from the asymmetric cost function perspective by combining diffusion strategy and the linear-linear cost (LLC), quadratic-quadratic cost (QQC), and linear-exponential cost (LEC), at all distributed network nodes, and named diffusion LLCLMS (DLLCLMS), diffusion QQCLMS (DQQCLMS), and diffusion LECLMS (DLECLMS), respectively. Then the stability of mean estimation error and computational complexity of those three diffusion algorithms are analyzed theoretically. Finally, several experiment simulation results are designed to verify the superiority of those three proposed diffusion algorithms. Experimental simulation results show that DLLCLMS, DQQCLMS, and DLECLMS algorithms are more robust to the input signal and impulsive noise than the DSELMS, DRVSSLMS, and DLLAD algorithms. In brief, theoretical analysis and experiment results show that those proposed DLLCLMS, DQQCLMS, and DLECLMS algorithms have superior performance when estimating the unknown linear system under the changeable impulsive noise environments and different types of input signals.

**Keywords**: Asymmetric cost function; adaptive diffusion algorithm; impulsive noise; input signals.

## 1 Introduction

So far, adaptive filter algorithms are often used in channel equalization, active interference control, echo cancellation, biomedical engineering [1-6], and many other fields. In recent years, especially the research on wireless sensor networks, in other words, diffusion adaptive filtering algorithms have been widely studied due to their unique performance, which is an extension of adaptive filtering algorithm over network

---

✉ Corresponding author;

Sihai Guan: gcihey@sina.cn;

This work was supported by the Fundamental Research Funds for the Central Universities Southwest Minzu University (Grant Number: 2021NQNCZ04) and the Wuhu and Xidian University special fund for industry-university-research cooperation (Project No: XWYCXY-012020014).

graphs [3]. Besides, there three collaborative strategies for adaptive filtering/estimation algorithms on the distributed network are widely used, including incremental strategies, consensus strategies, and diffusion strategies. But these three collaborative strategies have different performances; specifically, the consensus technique has an asymmetry problem, which can cause unstable growth. The diffusion strategies show good performance for unstable, and they can remove the asymmetry problem and real-time adaptation learning over distributed networks. So, the diffusion strategies were used frequently in the past decade, and they also include the adapt-then-combine (ATC) scheme [7] and the combine-then-adapt (CTA) scheme [3, 8]. Of course, the performance between these two schemes is also different; in response to this comparison, Cattivelli and colleagues analyzed these two schemes' performance, indicating that the ATC is better than the CTA [7]. With this conclusion, so in the following researches, the ATC scheme becomes a research focus in distributed adaptive filtering algorithms [7, 9-16].

The Wiener filter principle is the fundamental of the adaptive filtering algorithm based on the minimum mean square estimation error to construct an efficient convex cost function [17]. Obviously, we can know the core position of the cost function in the design of adaptive filtering algorithms. Besides, most of the cost functions were designed in the adaptive filtering algorithm field to satisfy symmetry, such as LMS [17], LMF [18], symmetric Gaussian kernel function [5], and hyperbolic function [19, 20]. However, symmetry and asymmetry are complementary concepts, so does the estimation error change symmetrical when running an adaptive filtering algorithm? Can an asymmetric function be used to design a cost function in an adaptive filtering algorithm? It is a problem that needs urgent attention to ensure the effectiveness of adaptive filtering algorithms. Usually, the change of the estimation error cost function will be affected by measurement interference/noise. The distribution of most interference or noise does not satisfy symmetry. So, for asymmetric estimation error distribution, the symmetric cost function is inappropriate and cannot adapt to the error distribution well. Asymmetric cost function usually performs very well, especially when the estimated error variable is of symmetric distribution. Therefore, for these asymmetric interferences, it is a feasible direction to use the asymmetric function to design the cost function to construct a novel adaptive filtering algorithm.

Among the asymmetric interference/noise, impulsive noise is the most representative asymmetrical distribution. Impulsive noise will especially significantly affect adaptive estimation accuracy and most diffusion adaptive estimation/filtering algorithms over a network graph. Therefore, it is necessary to design a robust distributed filtering algorithm for impulsive noise. For this purpose, there have been many papers so far [21-27]. The diffusion least mean p-power (DLMP) algorithm was proposed by Wen [21], which is robust to the generalized Gaussian noise distribution environments and prior knowledge of the distribution. However, the DLMP algorithm was proposed with a fixed power p-value, so parameter p-value is the critical factor, which means the DLMP algorithm performance is highly susceptible to the p-value. Based on the minimization of L1-norm subject to a constraint on the adaptive estimate vectors, Ni and colleagues designed a diffusion sign subband adaptive filtering (DSSAF) algorithm [22]. The DSSAF algorithm has better performance, but the DSSAF algorithm's computational complexity is relatively large. Besides, by combining the diffusion least mean square (DLMS) algorithm [10] and the sign operation to the estimated error at each iteration moment point, Ni and colleagues derived a diffusion sign-error LMS (DSELMS) algorithm [23]. The DSELMS algorithm has a simple architecture, but the DSELMS algorithm has a significant drawback, i.e., the steady-state estimation error is high [28]. Based on the Huber cost function, a similar set of diffusion adaptive filtering algorithms by Guan and colleagues [24], Wei and colleagues [25], and Soheila and colleagues [26] have been proposed as the DNHuber, DRVSSLMS, and RDLMS algorithms, respectively. But among them, the RDLMS algorithm is not practical for impulsive noise. The discussion of impulsive noise and input signal can be more comprehensive for the DNHuber algorithm. The DRVSSLMS algorithm has high algorithm computational complexity not conducive to implementing practical engineering. Besides, inspired by the least mean logarithmic absolute difference (LLAD) operation, Chen and colleagues designed another distributed adaptive filtering algorithm, i.e., the DLLAD algorithm [8]. But analysis of this distributed algorithm's robustness to the input signal and impulsive noise has not been performed. To solve this problem, Guan and colleagues proposed a diffusion probabilistic least mean square (DPLMS) algorithm [27] by combining the ATC scheme and the probabilistic LMS algorithm [29, 30] at all distributed network nodes.

Combining the content of the second and third paragraphs, we can see that most of the above distributed adaptive filtering algorithms are proposed by using the symmetry cost functions [31]. So, to address the asymmetry estimated error distribution issue, we propose a family diffusion adaptive filtering algorithms by using three different asymmetric cost functions (i.e., the linear-linear cost (LLC), quadratic-quadratic cost (QQC), and linear-exponential cost (LEC) [32-34]), namely the DLLCLMS, DQQCLMS, and DLECLMS algorithms in this paper. The stability of the mean estimation error and the computational complexity of those three proposed diffusion algorithms are analyzed theoretically. Simulation experiment results indicate that the DLLCLMS, DQQCLMS, and DLECLMS algorithms are more robust to the input signal and impulsive noise than the DSELMS, DRVSSLMS, and DLLAD algorithms.

The rest parts of this article are organized as follows. The proposed DLLCLMS, DQQCLMS, and DLECLMS algorithms will be described in detail in Section 2. The statistical stability behavior, computation complexity, and parameters ($a$ and $b$) of DLLCLMS, DQQCLMS, and DLECLMS algorithms are studied in Section 3. The simulation experiment is described in Section 4. Finally, conclusions are provided in

Section 5.

## 2 Proposed the diffusion algorithms using the asymmetric function

This section mainly describes in detail how to design the DLLCLMS, DQQCLMS, and DLECLMS algorithms. The specific plan is that the first step is to propose three adaptive filtering algorithms based on asymmetric cost functions include the LLC, QQC, and LEC functions, and the second step is to modify those three adaptive filtering algorithms by extending at all distributed network agents to propose the DLLCLMS, DQQCLMS, and DLECLMS algorithms.

### 2.1 Three adaptive filtering algorithms based on asymmetric cost functions

Setting an unknown linear system with the length $M$ of the system coefficient $\mathbf{W}^{\mathrm{o}}$, and $\mathbf{W}(i)$ be the adaptive estimated weight vector at iteration $i$, $\mathbf{X}(i)$ denote the input signal vector of the adaptive filtering algorithm and the estimation error $e(i)$ between the desired signal $d(i)$ and the estimation output $y(i)$ can be expressed as Eq. (1) ~Eq.(2). In addition, $v(i)$ is this unknown linear system measurement noise.

$$\begin{cases} d(i) = \mathbf{W}^{\mathrm{oT}}\mathbf{X}(i) + v(i) \\ \quad y(i) = \mathbf{W}^{\mathrm{T}}(i)\mathbf{X}(i) \end{cases} \tag{1}$$

$$e(i) = d(i) - y(i) = \mathbf{W}^{\mathrm{o}}\mathbf{X}(i) + v(i) - \mathbf{W}^{\mathrm{T}}(i)\mathbf{X}(i) \tag{2}$$

Next, three asymmetric cost functions include the LLC, QQC, and LEC functions, are used to design three adaptive filtering algorithms.

Firstly, LLC adaptive filtering algorithm aims to minimize the LLC cost function of estimation error defined as

$$J_{LLC}(i) = \begin{cases} a|e(i)|, if \;\; e(i) > 0 \\ b|e(i)|, if \;\; e(i) \le 0 \end{cases} \tag{3}$$

Secondly, QQC adaptive filtering algorithm aims to minimize the QQC cost function of estimation error defined as

$$J_{QQC}(i) = \frac{1}{2}\begin{cases} ae^2(i), if \;\; e(i) > 0 \\ be^2(i), if \;\; e(i) \le 0 \end{cases} \tag{4}$$

Thirdly, LEC adaptive filtering algorithm aims to minimize the LEC cost function of estimation error defined as

$$J_{LEC}(i) = b\left[\exp\left(ae(i)\right) - ae(i) - 1\right] \tag{5}$$

In Eq. (3), Eq. (4), and Eq. (5), $a, b > 0$ is the cut-off value.

Parameters $a$ and $b$ determine the shape and characteristics of each cost function; how to set it is very critical. Parameters $a$ and $b$ facilitate the adjustment of the asymmetric cost of error functions to the empirical cost situation because they determine the severity of a given estimation error type. For example, setting $a$=$b$ reduces QQC to the mean squared error, whereas LLC is reduced to the mean absolute error. From Eq. (3), the LLC cost function behaves as a sign-error cost function estimator. Therefore, the LLC cost function can combine the sign-error cost function and asymmetric estimated error. From Eq. (4), the QQC cost function behaves as a square error cost function estimator. Therefore, the QQC cost function can combine the square error cost function estimator and asymmetric estimated error. From Eq. (5), the LEC cost function behaves as an exponential function estimator. Therefore, the LEC cost function can combine exponential function estimator and asymmetric estimated error. From the above analysis, it can be seen that these three cost functions are worthy of further study and then used to design adaptive filtering algorithms.

According to the steepest descent method, the weight vector update of the LLC adaptive filter algorithm is

$$\mathbf{W}(i+1) = \begin{cases} \mathbf{W}(i) + \mu a sign\big(e(i)\big)\mathbf{X}(i), if \;\; e(i) > 0 \\ \mathbf{W}(i) + \mu b sign\big(e(i)\big)\mathbf{X}(i), if \;\; e(i) \le 0 \end{cases}$$
$$= \mathbf{W}(i) + \frac{\mu}{2}\left[a\left(1 + sign\big(e(i)\big)\right) + b\left(1 - sign\big(e(i)\big)\right)\right]\mathbf{X}(i) \tag{6}$$

In Eq. (6), $sign(\cdot)$ denotes the sign function, and $\mu$ is the step size.

According to the steepest descent method, the weight vector update of the QQC adaptive filter algorithm is

$$\mathbf{W}(i+1) = \begin{cases} \mathbf{W}(i) + \mu a e(i)\mathbf{X}(i), if \;\; e(i) > 0 \\ \mathbf{W}(i) + \mu b e(i)\mathbf{X}(i), if \;\; e(i) \le 0 \end{cases} \tag{7}$$

Similar operations, according to the steepest descent method, the weight vector update of the LEC adaptive filter algorithm is

$$\mathbf{W}(i+1) = \mathbf{W}(i) + \mu ab\left(\left(\exp\big(ae(i)\big) - 1\right)\right)\mathbf{X}(i) \tag{8}$$

In Eq. (8), the symbol $\exp(\cdot)$ denotes the exponential function, and $\mu$ is the step size.

### 2.2 Three asymmetric adaptive diffusion filtering algorithms

As described in the Introduction part, recently, the research on wireless sensor networks has been widely studied due to their unique performance. So, according to the design results of the previous subsection, three adaptive filtering algorithms, combined with a schematic diagram of distributed network structure in our previous research papers [24, 27], setting a distributed network of $N$ agent sensor nodes (as Fig.1). $\mathbf{X}_n(i)$ and $d_n(i)$ are the input signals and estimation output signals at agent $n$, respectively. It needs to be stated that this paper is different from our previous work [24, 27]. The similarity between them is how to realize the practical distributed adaptive estimation in impulsive interference. The difference is this paper using three asymmetric functions to design three cost functions to construct a novel family adaptive filtering algorithm to address the asymmetry estimated error distribution issue. And explore the robustness of the algorithm designed in this paper to the input signal and impulsive interference.

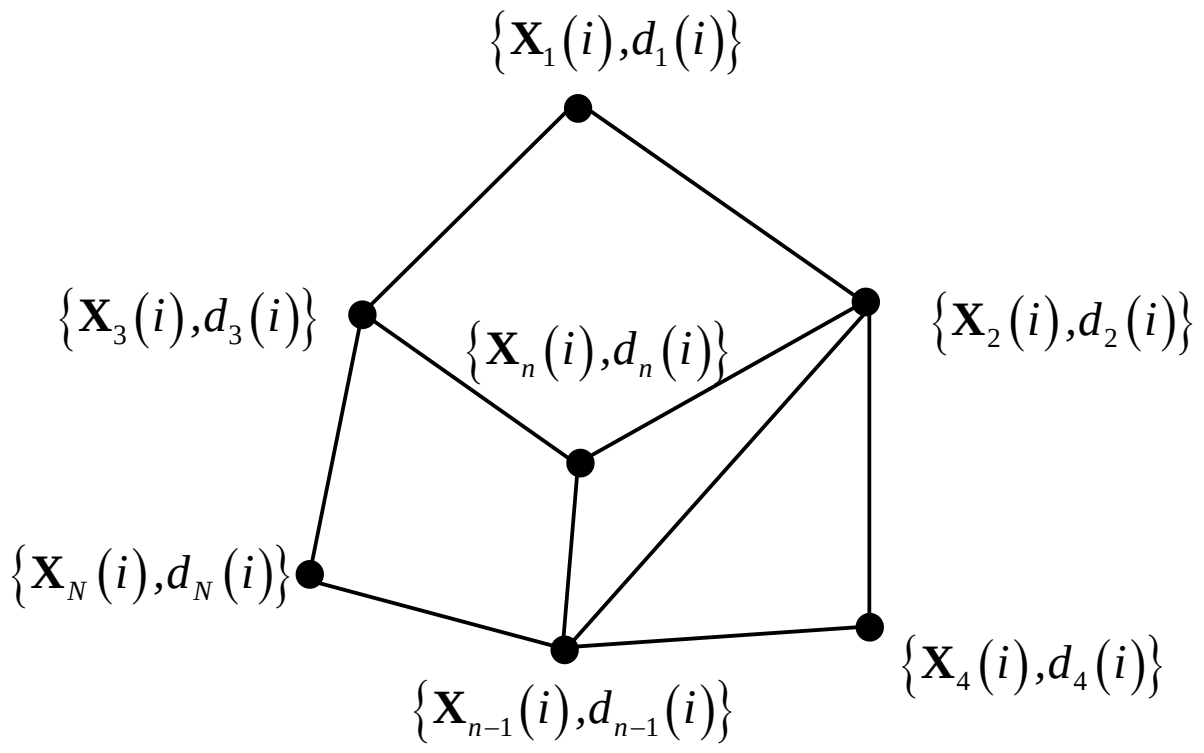

Fig.1 A distributed network consisting of *N* agent sensor nodes [24, 27].

Based on Fig. 1, by using minimizes the global cost function, we can seek the optimal linear estimator at each time instant *i*:

$$J^{global}(\mathbf{W}(i)) = \sum_n J_n^{local}(\mathbf{W}(i)) \tag{12}$$

where each sensor node $n \in \{1,2,\cdots,N\}$ has access to some zero-mean random process $\{d_n(i), \mathbf{X}_n(i)\}$, $d_n(i)$ is a scalar and $\mathbf{X}_n(i)$ is a regression vector. Suppose these measurement signals follow a standard computational model given by:

$$d_n(i) = \mathbf{W}^{\mathrm{oT}}\mathbf{X}_n(i) + v_n(i) \tag{13}$$

where $\mathbf{W}^{\mathrm{o}}$ is the unknown parameter vector with length *M*, and $v_n(i)$ is the unknown linear distributed network system measurement noise with variance $\sigma_{v,n}^2$.

The DLMS algorithm [10] is obtained by minimizing a linear combination of the local mean square estimation error:

$$J_n^{local}(\mathbf{W}(i)) = \sum_{l\in N_n} c_{l,n}\mathrm{E}\left[\left(e_l(i)\right)^2\right] = \sum_{l\in N_n} c_{l,n}\mathrm{E}\left[\left(d_l(i) - \mathbf{X}_l(i)\mathbf{W}^{\mathrm{T}}(i)\right)^2\right] \tag{14}$$

where the set of distributed network nodes that are connected to *n-th* node (including *n-th* node itself) is denoted by $N_n$ and is called the neighborhood of distributed network nodes *n*. The weighting coefficients $c_{l,n}$ are real and satisfy $\sum_{l\in N_n} c_{l,n} = 1$. $c_{l,n}$ forms a nonnegative combination matrix ***C***.

Cattivelli and colleagues analyzed ATC and CTA, and the ATC is better than the CTA [7]. So, using the ATC scheme, there are two steps in the DLMS algorithm: adaptation and combination. According to the order of these two steps, as follows

$$\begin{cases} \boldsymbol{\varphi}_n(i) = \mathbf{W}_n(i-1) + \mu_n\mathbf{X}_n(i)e_n(i) \\ \mathbf{W}_n(i) = \sum_{l\in N_n} c_{l,n}\boldsymbol{\varphi}_l(i) \end{cases} \tag{15}$$

where $\mu$ is the step-size (learning rate), and $\boldsymbol{\varphi}_n(i)$ is the local estimates at distributed network node *n*.

They were combining Eq. (6) ~Eq. (8) and Eq. (15), three asymmetric adaptive diffusion filtering algorithms are designed, as follows.

### (1) the DLLCLMS algorithm

Combining Eq. (6) and Eq. (15), a summary of the DLLCLMS algorithm procedure based on the analysis presented above is given in Table 1. From Table 1, we can know the DLLCLMS algorithm can be regarded as a general algorithm structure of the DSELMS algorithm. If $a = b$, the DLLCLMS algorithm is the DSELMS algorithm. In other words, the DLLCLMS can be seen as a mixture of a DSELMS algorithm for different estimated error $e(i)$ at a different network node and dynamic switching according to the relationship between $e(i)$ and 0.

Table 1 the DLLCLMS algorithm summary

Initialize: $\{\boldsymbol{w}_{n,-1} = \mathbf{0}\}$ for all agent $n$, $a$, $b$.

Set nonnegative combination weights $c_{l,n}$ for each time $i \geq 0$ and each agent $n$, and repeat:

$$\boldsymbol{\varphi}_n(i) = \begin{cases} \mathbf{W}_n(i-1) + \mu_n a sign(e_n(i))\mathbf{X}_n(i), if \;\; e_n(i) > 0 & (I) \\ \mathbf{W}_n(i-1) + \mu_n b sign(e_n(i))\mathbf{X}_n(i), if \;\; e_n(i) \leq b & (II) \end{cases} \tag{16}$$

$$\mathbf{W}_n(i) = \sum_{l \in N_n} c_{l,n} \boldsymbol{\varphi}_l(i) \tag{17}$$

**(2) the DQQCLMS algorithm**

Combining Eq. (7) and Eq. (15), a summary of the DQQCLMS algorithm procedure based on the analysis presented above is given in Table 2. From Table 2, we can know the DLLCLMS algorithm can be regarded as a general algorithm structure of the DLMS algorithm. If $a = b$, the DLLCLMS algorithm is the DLMS algorithm. In other words, the DLLCLMS can be seen as a mixture of a DLMS algorithm for different estimated error $e(i)$ at a different network node and dynamic switching according to the relationship between $e(i)$ and 0.

Table 2 the DQQCLMS algorithm summary

Initialize: $\{\boldsymbol{w}_{n,-1} = \mathbf{0}\}$ for all agent $n$, $a$, $b$.

Set nonnegative combination weights $c_{l,n}$ for each time $i \geq 0$ and each agent $n$, and repeat:

$$\boldsymbol{\varphi}_n(i) = \begin{cases} \mathbf{W}_n(i-1) + \mu_n a e_n(i)\mathbf{X}_n(i), if \;\; e_n(i) > 0 & (I) \\ \mathbf{W}_n(i-1) + \mu_n b e_n(i)\mathbf{X}_n(i), if \;\; e_n(i) \leq b & (II) \end{cases} \tag{18}$$

$$\mathbf{W}_n(i) = \sum_{l \in N_n} c_{l,n} \boldsymbol{\varphi}_l(i) \tag{19}$$

**(3) the DLECLMS algorithm**

Combining Eq. (8) and Eq. (15), a summary of the DLECLMS algorithm procedure based on the analysis presented above is given in Table 3. Table 3 shows that the DLECLMS algorithm can be regarded as an adaptive filter based on an asymmetric exponential function estimator and extend in a distributed network.

Table 3 the DLECLMS algorithm summary

Initialize: $\{\boldsymbol{w}_{n,-1} = \mathbf{0}\}$ for all agent $n$, $a$, $b$.

Set nonnegative combination weights $c_{l,n}$ for each time $i \geq 0$ and each agent $n$, and repeat:

$$\boldsymbol{\varphi}_n(i) = \mathbf{W}_n(i-1) + \mu_n ab\left(\left(\exp(a e_n(i)) - 1\right)\right)\mathbf{X}_n(i) \tag{20}$$

$$\mathbf{W}_n(i) = \sum_{l \in N_n} c_{l,n} \boldsymbol{\varphi}_l(i) \tag{21}$$

## 3. Performance of proposed diffusion algorithms

After completing those three diffusion adaptive filtering algorithms design, the performance of those three algorithms should be analyzed theoretically. This subsection will discuss the performances of the diffusion asymmetric adaptive filtering algorithms, including mean behavior and computational complexity.

To facilitate performance analysis, we make the following assumptions:

*Assumption 1: The distributed network system measurement noises are independent of any other signals.*

*Assumption 2:* $\boldsymbol{X}(i)$ *is zero-mean Gaussian, temporally white, and spatially independent with* $\boldsymbol{R}_{xx,n} = E[\boldsymbol{X_n}(i)\boldsymbol{X_n}^T(i)]$.

*Assumption 3: The regression vector* $\boldsymbol{X}_n(i)$ *is independent of* $\widehat{\boldsymbol{W}}_n(j)$ *for all distributed networks n and j < i. All distributed network system weight vectors are approximately independent of all input signals.*

*Assumption 4: The distributed network system measurement noises* $v_n(i)$ *at the n-th agent is assumed to be a mixture signal of zero-mean white Gaussian noise* $g_n(i)$ *of variance* $\sigma_{g,n}^2$ *and impulsive noise* $Im_n(i)$*, i.e.,* $v_n(i) = g_n(i) + Im_n(i)$*. The impulsive noise can be described by using the equation as* $Im_n(i) = B_n(i)G_n(i)$*, where* $B_n(i)$ *is a Bernoulli process with the probability of* $P[B_n(i) = 1] = P_r$ *and* $P[B_n(i) = 0] = 1 - P_r$*, and* $G_n(i)$ *is a zero-mean white Gaussian process of variance* $I_n\sigma_{g,n}^2$ *with* $I_n \gg 1$.

Then, let us define some equations at agent *n* and time *i*, $\widehat{\mathbf{W}}_n(i) = \mathbf{W}^o - \mathbf{W}_n(i)$, $\widehat{\boldsymbol{\varphi}}_n(i) = \mathbf{W}^o - \boldsymbol{\varphi}_n(i)$, which are then collected to form the network weight error vector and intermediate network weight error vector, i.e., $\mathbf{W}(i) = \mathrm{col}\{\mathbf{W}_1(i), \mathbf{W}_2(i), \cdots, \mathbf{W}_N(i)\}$, $\boldsymbol{\varphi}(i) = \mathrm{col}\{\boldsymbol{\varphi}_1(i), \boldsymbol{\varphi}_2(i), \cdots, \boldsymbol{\varphi}_N(i)\}$, $\widehat{\mathbf{W}}(i) = \mathrm{col}\{\widehat{\mathbf{W}}_1(i), \widehat{\mathbf{W}}_2(i), \cdots, \widehat{\mathbf{W}}_N(i)\}$, $\widehat{\boldsymbol{\varphi}}(i) = \mathrm{col}\{\widehat{\boldsymbol{\varphi}}_1(i), \widehat{\boldsymbol{\varphi}}_2(i), \cdots, \widehat{\boldsymbol{\varphi}}_N(i)\}$, $\boldsymbol{\mu_A} = \mathrm{diag}\{a\mu_1, a\mu_2, \cdots, a\mu_N\}$, $\boldsymbol{\mu_B} = \mathrm{diag}\{b\mu_1, b\mu_2, \cdots, b\mu_N\}$, $\boldsymbol{\mu_{AB}} = \mathrm{diag}\{ab\mu_1, ab\mu_2, \cdots, ab\mu_N\}$, and $\mathbf{e}(i) = \mathrm{col}\{e_1(i), e_2(i), \cdots, e_N(i)\}$.

## 3.1 Mean weight vector error behavior

Two noteworthy performances of adaptive filtering algorithms are convergence and steady-state characteristics. So, by studying the mean weight estimation error vector, the convergence and steady-state error properties of those three proposed diffusion adaptive filtering algorithms can be explored. The following will analyze these three diffusion algorithms' mean weight vector error behavior performance one by one.

### (1) the DLLCLMS algorithm

Eq. (16) and Eq. (17) can be written as

$$\widehat{\boldsymbol{\varphi}}(i) = \begin{cases} \widehat{\mathbf{W}}(i-1) - \boldsymbol{S}_{\mu_A}\mathbf{S}_{\mathbf{S}_e}(i)\mathbf{X}(i), if \ \ \mathbf{e}(i) > \mathbf{0} & (I) \\ \widehat{\mathbf{W}}(i-1) - \boldsymbol{S}_{\mu_B}\mathbf{S}_{\mathbf{S}_e}(i)\mathbf{X}(i), if \ \ \mathbf{e}(i) \le \mathbf{0} & (II) \end{cases} \tag{22}$$

$$\widehat{\mathbf{W}}(i) = \mathbf{C}^T\widehat{\boldsymbol{\varphi}}(i) \tag{23}$$

where $\mathbf{C} = \mathbf{C} \otimes \mathbf{I}$, $\boldsymbol{S}_{\mu_A} = \boldsymbol{\mu_A} \otimes \mathbf{I}$, $\mathbf{S}_{\mathbf{S}_e}(i) = \mathbf{S}_e(i) \otimes \mathbf{I}$, $\mathbf{X}(i) = \mathrm{col}\{\mathbf{X}_1(i), \mathbf{X}_2(i), \cdots, \mathbf{X}_N(i)\}$, $\mathbf{S}_e(i) = \mathrm{diag}\{\mathrm{sign}(\mathbf{e}(i))\}$, and $\otimes$ denotes the Kronecker product operation.

Taking the expectation of Eq. (22) and Eq. (23),

$$\mathrm{E}[\widehat{\mathbf{W}}(i)] = \begin{cases} \mathbf{C}^{\mathrm{T}}\mathbf{E}[\widehat{\mathbf{W}}(i-1)] - \mathbf{C}^{\mathrm{T}}\boldsymbol{S}_{\mu_A}\mathbf{E}[\mathbf{S}_{\mathrm{S_e}}(i)\mathbf{X}(i)], if \ \ \mathbf{e}(i) > \mathbf{0} \quad (I) \\ \mathbf{C}^{\mathrm{T}}\mathbf{E}[\widehat{\mathbf{W}}(i-1)] - \mathbf{C}^{\mathrm{T}}\boldsymbol{S}_{\mu_B}\mathbf{E}[\mathbf{S}_{\mathrm{S_e}}(i)\mathbf{X}(i)], if \ \ \mathbf{e}(i) \le \mathbf{0} \quad (II) \end{cases} \tag{24}$$

Denote the measurement noise vector by $\mathbf{V}(i) = \mathrm{col}\{v_1(i), v_2(i), \cdots, v_N(i)\}$, $\mathbf{g}(i) = \mathrm{col}\{g_1(i), g_2(i), \cdots, g_N(i)\}$, $\mathbf{Im}(i) = \mathrm{col}\{Im_1(i), Im_2(i), \cdots, Im_N(i)\}$, $\mathbf{S_g}(i) = \mathrm{diag}\{\mathrm{sign}(\mathbf{g}(i))\}$, $\mathbf{S_{Im}}(i) = \mathrm{diag}\{\mathrm{sign}(\mathbf{Im}(i))\}$, $\mathbf{S}_{\boldsymbol{X}}(i) = \mathrm{diag}\{\mathbf{X}_1(i), \mathbf{X}_2(i), \cdots, \mathbf{X}_N(i)\}$. So, from Eq. (1), we have $\mathbf{e}(i) = \mathbf{S}_{\boldsymbol{X}}{}^{\mathrm{T}}(i)\widehat{\mathbf{W}}(i-1) + \mathbf{V}(i) = \mathbf{e}_{\boldsymbol{o}}(i) + \mathbf{V}(i)$.

Then, let $\begin{cases} \mathbf{e}_{\boldsymbol{g}}(i) = \mathbf{e}_{\boldsymbol{o}}(i) + \mathbf{g}(i) \\ \mathbf{e}_{\boldsymbol{Im}(i)} = \mathbf{e}_{\boldsymbol{o}}(i) + \mathbf{Im}(i) \end{cases}$, $\begin{cases} \boldsymbol{S}_{\mathbf{S}_g}(i) = \mathbf{S}_{\boldsymbol{g}}(i) \otimes \mathbf{I} \\ \boldsymbol{S}_{\mathbf{S}_{Im}}(i) = \mathbf{S}_{\boldsymbol{Im}}(i) \otimes \mathbf{I} \end{cases}$.

So,

$$\mathbf{E}[\mathbf{S}_{\mathrm{S_e}}(i)\mathbf{X}(i)] = (1-P_r)\mathbf{E}\left[\mathbf{S}_{\mathrm{S_g}}(i)\mathbf{X}(i)\right] + P_r\mathbf{E}[\boldsymbol{S}_{\mathbf{S}_{Im}}(i)\mathbf{X}(i)] \tag{25}$$

Let

$$\begin{cases} \sigma^2_{e_{g,n}}(i) = \mathrm{Tr}[\mathbf{R}_{ww,n}(i-1)\mathbf{R}_{xx,n}] + \sigma^2_{g,n}(i) \\ \sigma^2_{e_{Im,n}}(i) = \mathrm{Tr}[\mathbf{R}_{ww,n}(i-1)\mathbf{R}_{xx,n}] + \sigma^2_{Im,n}(i) \end{cases},$$

$$\begin{cases} \mathbf{X}_{\boldsymbol{g}}(i) = \sqrt{\frac{2}{\pi}}\mathrm{diag}\{\sigma^{-1}_{g,1}(i), \sigma^{-1}_{g,2}(i), \cdots, \sigma^{-1}_{g,N}(i)\} \\ \mathbf{X}_{\boldsymbol{Im}}(i) = \sqrt{\frac{2}{\pi}}\mathrm{diag}\{\sigma^{-1}_{Im,1}(i), \sigma^{-1}_{Im,2}(i), \cdots, \sigma^{-1}_{Im,N}(i)\} \end{cases},$$

$$\begin{cases} \boldsymbol{S}_{\mathrm{X}_g}(i) = \mathbf{X}_{\boldsymbol{g}}(i) \otimes \mathbf{I} \\ \boldsymbol{S}_{\mathrm{X}_{Im}}(i) = \mathbf{X}_{\boldsymbol{Im}}(i) \otimes \mathbf{I} \end{cases}.$$

Then

$$\begin{cases} \mathbf{E}\left[\mathbf{S}_{\mathrm{S_g}}(i)\mathbf{X}(i)|\mathbf{W}(i-1)\right] = \mathbf{X}_{\boldsymbol{g}}(i)\mathrm{diag}\{\mathbf{R}_{xx,1}, \mathbf{R}_{xx,2}, \cdots, \mathbf{R}_{xx,N}\}\widehat{\mathbf{W}}(i-1) \\ \mathbf{E}[\mathbf{S}_{\mathrm{S_{Im}}}(i)\mathbf{X}(i)|\mathbf{W}(i-1)] = \mathbf{X}_{\boldsymbol{Im}}(i)\mathrm{diag}\{\mathbf{R}_{xx,1}, \mathbf{R}_{xx,2}, \cdots, \mathbf{R}_{xx,N}\}\widehat{\mathbf{W}}(i-1) \end{cases} \tag{26}$$

Substituting (26) into (25), we have

$$\begin{aligned} &\mathbf{E}[\mathbf{S}_{\mathrm{S_e}}(i)\mathbf{X}(i)] \\ &\approx \mathbf{E}\left[\mathbf{E}[\mathbf{S}_{\mathrm{S_e}}(i)\mathbf{X}(i)|\mathbf{W}(i-1)]\right] \\ &= \left[(1-P_r)\mathbf{S}_{\mathrm{S_g}}(i) + P_r\mathbf{S}_{\mathrm{S_{Im}}}(i)\right]\mathrm{diag}\{\mathbf{R}_{xx,1}, \mathbf{R}_{xx,2}, \cdots, \mathbf{R}_{xx,N}\}\mathbf{E}[\widehat{\mathbf{W}}(i-1)] \end{aligned} \tag{27}$$

Finally, substituting (27) into (24) obtains

$$\mathrm{E}[\widehat{\mathbf{W}}(i)] = \begin{cases} \mathbf{C}^{\mathrm{T}}\left[\boldsymbol{I}_{\boldsymbol{NM}} - \boldsymbol{S}_{\mu_A}\left((1-P_r)\mathbf{S}_{\mathrm{S_g}}(i) + P_r\mathbf{S}_{\mathrm{S_{Im}}}(i)\right)\mathrm{diag}\{\mathbf{R}_{xx,1}, \mathbf{R}_{xx,2}, \cdots, \mathbf{R}_{xx,N}\}\right]\mathbf{E}[\widehat{\mathbf{W}}(i-1)], if \ \ \mathbf{e}(i) > \mathbf{0} \ \ (I) \\ \mathbf{C}^{\mathrm{T}}\left[\boldsymbol{I}_{\boldsymbol{NM}} - \boldsymbol{S}_{\mu_B}\left((1-P_r)\mathbf{S}_{\mathrm{S_g}}(i) + P_r\mathbf{S}_{\mathrm{S_{Im}}}(i)\right)\mathrm{diag}\{\mathbf{R}_{xx,1}, \mathbf{R}_{xx,2}, \cdots, \mathbf{R}_{xx,N}\}\right]\mathbf{E}[\widehat{\mathbf{W}}(i-1)], if \ \ \mathbf{e}(i) \le \mathbf{0} \ \ (II) \end{cases} \tag{28}$$

From Eq. (28), one can see that the asymptotic unbiasedness of the DLLCLMS algorithm can be guaranteed if the matrix $\mathbf{C}^{\mathrm{T}}\left[\boldsymbol{I}_{\boldsymbol{NM}} - \boldsymbol{S}_{\mu_A}\left((1-P_r)\mathbf{S}_{\mathrm{S_g}}(i) + P_r\mathbf{S}_{\mathrm{S_{Im}}}(i)\right)\mathrm{diag}\{\mathbf{R}_{xx,1}, \mathbf{R}_{xx,2}, \cdots, \mathbf{R}_{xx,N}\}\right]$ and $\mathbf{C}^{\mathrm{T}}\left[\boldsymbol{I}_{\boldsymbol{NM}} - \boldsymbol{S}_{\mu_B}\left((1-P_r)\mathbf{S}_{\mathrm{S_g}}(i) + P_r\mathbf{S}_{\mathrm{S_{Im}}}(i)\right)\mathrm{diag}\{\mathbf{R}_{xx,1}, \mathbf{R}_{xx,2}, \cdots, \mathbf{R}_{xx,N}\}\right]$ are stable. Both of the matrix $\left[\boldsymbol{I}_{\boldsymbol{NM}} - \boldsymbol{S}_{\mu_A}\left((1-P_r)\mathbf{S}_{\mathrm{S_g}}(i) + P_r\mathbf{S}_{\mathrm{S_{Im}}}(i)\right)\mathrm{diag}\{\mathbf{R}_{xx,1}, \mathbf{R}_{xx,2}, \cdots, \mathbf{R}_{xx,N}\}\right]$ and $\left[\boldsymbol{I}_{\boldsymbol{NM}} - \boldsymbol{S}_{\mu_B}\left((1-P_r)\mathbf{S}_{\mathrm{S_g}}(i) + \right.\right.$

$P_r\mathbf{S}_{\mathbf{S}_{\mathbf{Im}}}(i)\Big)\operatorname{diag}\{\mathbf{R}_{xx,1},\mathbf{R}_{xx,2},\cdots,\mathbf{R}_{xx,N}\}\Big]$ are a block-diagonal matrix and it can be easily verified that it is stable if its block-diagonal entries $[\boldsymbol{I}-a\mu_n \mathrm{X}_v(i)\mathbf{R}_{xx,n}]$ and $[\boldsymbol{I}-b\mu_n \mathrm{X}_v(i)\mathbf{R}_{xx,n}]$ are stable, where $\mathrm{X}_v(i)\leq\sqrt{\frac{2}{\pi}}\big[(1-P_r)\sigma_{g,1}^{-1}(i)+P_r\sigma_{Im,1}^{-1}(i)\big]$. So, the condition for stability of the mean weight error vector is given by

$$\begin{cases}0<\mu_n<\frac{2}{a\mathrm{X}_v(i)\rho_{max}(\mathbf{R}_{xx,n})},if\ \ e_n(i)>0\ \ (I)\\ 0<\mu_n<\frac{2}{b\mathrm{X}_v(i)\rho_{max}(\mathbf{R}_{xx,n})},if\ \ e_n(i)\leq 0\ \ (II)\end{cases} \tag{29}$$

where $\rho_{max}$ denotes the maximal eigenvalue of $\mathbf{R}_{xx,n}$. So, based on Eq. (29) and Eq. (24), we obtain $\mathrm{E}[\widehat{\mathbf{W}}(\infty)]=\mathbf{0}$.

**(2) the DQQCLMS algorithm**

Eq. (18) and Eq. (19) can be written as

$$\widehat{\boldsymbol{\varphi}}(i)=\begin{cases}\widehat{\mathbf{W}}(i-1)-\boldsymbol{S}_{\mu_A}\mathbf{S}_{\mathbf{S}_e}(i)\mathbf{X}(i),if\ \ \mathbf{e}(i)>\mathbf{0}\quad (I)\\ \widehat{\mathbf{W}}(i-1)-\boldsymbol{S}_{\mu_B}\mathbf{S}_{\mathbf{S}_e}(i)\mathbf{X}(i),if\ \ \mathbf{e}(i)\leq\mathbf{0}\quad (II)\end{cases} \tag{30}$$

$$\widehat{\mathbf{W}}(i)=\mathbf{C}^{\mathrm{T}}\widehat{\boldsymbol{\varphi}}(i) \tag{31}$$

where $\mathbf{C}=\mathbf{C}\otimes\mathbf{I}$, $\boldsymbol{S}_{\mu_A}=\boldsymbol{\mu}_A\otimes\mathbf{I}$, $\mathbf{S}_{\mathbf{S}_e}(i)=\mathbf{S}_e(i)\otimes\mathbf{I}$, $\mathbf{X}(i)=\operatorname{col}\{\mathbf{X}_1(i),\mathbf{X}_2(i),\cdots,\mathbf{X}_N(i)\}$, $\mathbf{S}_e(i)=\operatorname{diag}\{\mathbf{e}(i)\}$, and $\otimes$ denotes the Kronecker product operation.
Taking the expectation of Eq. (30) and Eq. (31),

$$\mathrm{E}\big[\widehat{\mathbf{W}}(i)\big]=\begin{cases}\mathbf{C}^{\mathrm{T}}\mathbf{E}\big[\widehat{\mathbf{W}}(i-1)\big]-\mathbf{C}^{\mathrm{T}}\boldsymbol{S}_{\mu_A}\mathbf{E}\big[\mathbf{S}_{\mathbf{S}_e}(i)\mathbf{X}(i)\big],if\ \ \mathbf{e}(i)>\mathbf{0}\quad (I)\\ \mathbf{C}^{\mathrm{T}}\mathbf{E}\big[\widehat{\mathbf{W}}(i-1)\big]-\mathbf{C}^{\mathrm{T}}\boldsymbol{S}_{\mu_B}\mathbf{E}\big[\mathbf{S}_{\mathbf{S}_e}(i)\mathbf{X}(i)\big],if\ \ \mathbf{e}(i)\leq\mathbf{0}\quad (II)\end{cases} \tag{32}$$

Denote the measurement noise vector by $\mathbf{V}(i)=\operatorname{col}\{v_1(i),v_2(i),\cdots,v_N(i)\}$, $\mathbf{g}(i)=\operatorname{col}\{g_1(i),g_2(i),\cdots,g_N(i)\}$, $\mathbf{Im}(i)=\operatorname{col}\{Im_1(i),Im_2(i),\cdots,Im_N(i)\}$, $\mathbf{S_g}(i)=\operatorname{diag}\{\mathbf{g}(i)\}$, $\mathbf{S_{Im}}(i)=\operatorname{diag}\{\mathbf{Im}(i)\}$, $\mathbf{S}_{\boldsymbol{X}}(i)=\operatorname{diag}\{\mathbf{X}_1(i),\mathbf{X}_2(i),\cdots,\mathbf{X}_N(i)\}$. So, from Eq. (1), we have $\mathbf{e}(i)=\mathbf{S}_{\boldsymbol{X}}{}^{\mathrm{T}}(i)\widehat{\mathbf{W}}(i-1)+\mathbf{V}(i)=\mathbf{e}_{\boldsymbol{o}}(i)+\mathbf{V}(i)$.

Then, let $\begin{cases}\mathbf{e}_{\boldsymbol{g}}(i)=\mathbf{e}_{\boldsymbol{o}}(i)+\mathbf{g}(i)\\ \mathbf{e}_{\boldsymbol{Im}(i)}=\mathbf{e}_{\boldsymbol{o}}(i)+\mathbf{Im}(i)\end{cases}$, $\begin{cases}\boldsymbol{S}_{\mathbf{S}_g}(i)=\mathbf{S}_{\boldsymbol{g}}(i)\otimes\mathbf{I}\\ \boldsymbol{S}_{\mathbf{S}_{Im}}(i)=\mathbf{S}_{\boldsymbol{Im}}(i)\otimes\mathbf{I}\end{cases}$.

So,

$$\mathrm{E}\big[\widehat{\mathbf{W}}(i)\big]=\begin{cases}\mathbf{C}^{\mathrm{T}}\big[\boldsymbol{I}_{\boldsymbol{NM}}-\boldsymbol{S}_{\mu_A}\operatorname{diag}\{\mathbf{R}_{xx,1},\mathbf{R}_{xx,2},\cdots,\mathbf{R}_{xx,N}\}\big]\mathbf{E}\big[\widehat{\mathbf{W}}(i-1)\big],if\ \ \mathbf{e}(i)>\mathbf{0}\ \ (I)\\ \mathbf{C}^{\mathrm{T}}\big[\boldsymbol{I}_{\boldsymbol{NM}}-\boldsymbol{S}_{\mu_B}\operatorname{diag}\{\mathbf{R}_{xx,1},\mathbf{R}_{xx,2},\cdots,\mathbf{R}_{xx,N}\}\big]\mathbf{E}\big[\widehat{\mathbf{W}}(i-1)\big],if\ \ \mathbf{e}(i)\leq\mathbf{0}\ \ (II)\end{cases} \tag{33}$$

From Eq. (33), one can see that the asymptotic unbiasedness of the DQQCLMS algorithm can be guaranteed if the matrix $\mathbf{C}^{\mathrm{T}}\big[\boldsymbol{I}_{\boldsymbol{NM}}-\boldsymbol{S}_{\mu_A}\operatorname{diag}\{\mathbf{R}_{xx,1},\mathbf{R}_{xx,2},\cdots,\mathbf{R}_{xx,N}\}\big]$ and $\mathbf{C}^{\mathrm{T}}\big[\boldsymbol{I}_{\boldsymbol{NM}}-\boldsymbol{S}_{\mu_B}\operatorname{diag}\{\mathbf{R}_{xx,1},\mathbf{R}_{xx,2},\cdots,\mathbf{R}_{xx,N}\}\big]$ are stable. Both of the matrix $\big[\boldsymbol{I}_{\boldsymbol{NM}}-\boldsymbol{S}_{\mu_A}\operatorname{diag}\{\mathbf{R}_{xx,1},\mathbf{R}_{xx,2},\cdots,\mathbf{R}_{xx,N}\}\big]$ and $\big[\boldsymbol{I}_{\boldsymbol{NM}}-\boldsymbol{S}_{\mu_B}\operatorname{diag}\{\mathbf{R}_{xx,1},\mathbf{R}_{xx,2},\cdots,\mathbf{R}_{xx,N}\}\big]$ are a block-diagonal matrix and it can be easily verified that it is stable if its block-diagonal entries $[\boldsymbol{I}-a\mu_n\mathbf{R}_{xx,n}]$ and $[\boldsymbol{I}-b\mu_n\mathbf{R}_{xx,n}]$ are stable. So, the condition for stability of the mean weight error vector is given by

$$\begin{cases}0<\mu_n<\frac{2}{a\rho_{max}(\mathbf{R}_{xx,n})},if\ \ e_n(i)>0\ \ (I)\\ 0<\mu_n<\frac{2}{b\rho_{max}(\mathbf{R}_{xx,n})},if\ \ e_n(i)\leq 0\ \ (II)\end{cases} \tag{34}$$

where $\rho_{max}$ denotes the maximal eigenvalue of $\mathbf{R}_{xx,n}$. So, based on Eq. (32) and Eq.

(34), we obtain $\mathrm{E}[\widehat{\mathbf{W}}(\infty)] = \mathbf{0}$.

**(3) the DLECLMS algorithm**

Based on Eq.(20), let $J_{f_{DLECLMS}}(i) = \exp(ae_n(i)) - 1$, then this can be observed from the Taylor series expansion of $J_{f_{DLECLMS}}(i)$ around $e_n(i) = 0$,

$$J_{f_{DLECLMS}}(i) = \exp(ae_n(i)) - 1 = a\sum_{k=1}^{+\infty}\frac{1}{k!}e_n^k(i) \tag{35}$$

As desired, since the weight of the $k$ −th error moment is $\frac{1}{k!}$ , it is given to lower-order moments. Notice also that for small error values, the error cost functions become,

$$J_{f_{DLECLMS}}(i) = \exp(ae_n(i)) - 1 = a\sum_{k=1}^{+\infty}\frac{1}{k!}e_n^k(i) \approx ae_n(i) \tag{36}$$

Eq. (20) and Eq. (21) can be written as

$$\widehat{\boldsymbol{\varphi}}(i) = \widehat{\mathbf{W}}(i-1) - \boldsymbol{S}_{\boldsymbol{\mu}_{AB}}\mathbf{S}_{\mathbf{S}_{\mathrm{e}}}(i)\mathbf{X}(i) \tag{37}$$

$$\widehat{\mathbf{W}}(i) = \mathbf{C}^{\mathrm{T}}\widehat{\boldsymbol{\varphi}}(i) \tag{38}$$

where $\mathbf{C} = \mathbf{C} \otimes \mathbf{I}$, $\boldsymbol{S}_{\boldsymbol{\mu}_{AB}} = \boldsymbol{\mu}_{AB} \otimes \mathbf{I}$, $\mathbf{S}_{\mathbf{S}_{\mathrm{e}}}(i) = \mathbf{S}_{\mathrm{e}}(i) \otimes \mathbf{I}$, $\mathbf{X}(i) = \mathrm{col}\{\mathbf{X}_1(i), \mathbf{X}_2(i), \cdots, \mathbf{X}_N(i)\}$, $\mathbf{S}_{\mathrm{e}}(i) = \mathrm{diag}\{\exp(a\mathbf{e}(i)) - 1\} \approx \mathrm{diag}\{a\mathbf{e}(i)\}$, and $\otimes$ denotes the Kronecker product operation.

Taking the expectation of Eq. (37) and Eq. (38),

$$\mathrm{E}[\widehat{\mathbf{W}}(i)] = \mathbf{C}^{\mathrm{T}}\mathbf{E}[\widehat{\mathbf{W}}(i-1)] - \mathbf{C}^{\mathrm{T}}\boldsymbol{S}_{\boldsymbol{\mu}_{AB}}\mathbf{E}[\mathbf{S}_{\mathbf{S}_{\mathrm{e}}}(i)\mathbf{X}(i)] \tag{39}$$

Denote the measurement noise vector by $\mathbf{V}(i) = \mathrm{col}\{v_1(i), v_2(i), \cdots, v_N(i)\}$, $\mathbf{g}(i) = \mathrm{col}\{g_1(i), g_2(i), \cdots, g_N(i)\}$, $\mathbf{Im}(i) = \mathrm{col}\{Im_1(i), Im_2(i), \cdots, Im_N(i)\}$, $\mathbf{S}_{\mathbf{g}}(i) = \mathrm{diag}\{\mathbf{g}(i)\}$, $\mathbf{S}_{\mathbf{Im}}(i) = \mathrm{diag}\{\mathbf{Im}(i)\}$, $\mathbf{S}_{\boldsymbol{X}}(i) = \mathrm{diag}\{\mathbf{X}_1(i), \mathbf{X}_2(i), \cdots, \mathbf{X}_N(i)\}$. So, from Eq. (1), we have $\mathbf{e}(i) = \mathbf{S}_{\boldsymbol{X}}{}^{\mathrm{T}}(i)\widehat{\mathbf{W}}(i-1) + \mathbf{V}(i) = \mathbf{e}_{\boldsymbol{o}}(i) + \mathbf{V}(i)$.

Then, let $\begin{cases} \mathbf{e}_{\boldsymbol{g}}(i) = \mathbf{e}_{\boldsymbol{o}}(i) + \mathbf{g}(i) \\ \mathbf{e}_{\boldsymbol{Im}(i)} = \mathbf{e}_{\boldsymbol{o}}(i) + \mathbf{Im}(i) \end{cases}$, $\begin{cases} \boldsymbol{S}_{\mathbf{S}_{\boldsymbol{g}}}(i) = \mathbf{S}_{\boldsymbol{g}}(i) \otimes \mathbf{I} \\ \boldsymbol{S}_{\mathbf{S}_{\boldsymbol{Im}}}(i) = \mathbf{S}_{\boldsymbol{Im}}(i) \otimes \mathbf{I} \end{cases}$.

So,

$$\mathrm{E}[\widehat{\mathbf{W}}(i)] = \mathbf{C}^{\mathrm{T}}[\boldsymbol{I}_{\boldsymbol{NM}} - \boldsymbol{S}_{\boldsymbol{\mu}_{AB}}\mathrm{diag}\{a\mathbf{R}_{xx,1}, a\mathbf{R}_{xx,2}, \cdots, a\mathbf{R}_{xx,N}\}]\mathbf{E}[\widehat{\mathbf{W}}(i-1)] \tag{40}$$

From Eq. (40), one can see that the asymptotic unbiasedness of the DLECLMS algorithm can be guaranteed if the matrix $\mathbf{C}^{\mathrm{T}}[\boldsymbol{I}_{\boldsymbol{NM}} - \boldsymbol{S}_{\boldsymbol{\mu}_{AB}}\mathrm{diag}\{a\mathbf{R}_{xx,1}, a\mathbf{R}_{xx,2}, \cdots, a\mathbf{R}_{xx,N}\}]$ is stable. The matrix $[\boldsymbol{I}_{\boldsymbol{NM}} - \boldsymbol{S}_{\boldsymbol{\mu}_{AB}}\mathrm{diag}\{a\mathbf{R}_{xx,1}, a\mathbf{R}_{xx,2}, \cdots, a\mathbf{R}_{xx,N}\}]$ is a block-diagonal matrix and it can be easily verified that it is stable if its block-diagonal entries $[\boldsymbol{I} - a^2 b\mu_n \mathbf{R}_{xx,n}]$ are stable. So, the condition for stability of the mean weight error vector (as Eq. (40)) is given by

$$0 < \mu_n < \frac{2}{a^2 b\rho_{max}(\mathbf{R}_{xx,n})} \tag{41}$$

where $\rho_{max}$ denotes the maximal eigenvalue of $\mathbf{R}_{xx,n}$. So, based on Eq. (40) and Eq. (41), we obtain $\mathrm{E}[\widehat{\mathbf{W}}(\infty)] = \mathbf{0}$.

## 3.2 Computational complexity

Another important indicator to measure the performance of an adaptive filtering algorithm is the computational complexity because it directly determines whether the adaptive filtering algorithm is easy to implement in engineering. The diffusion adaptive filtering algorithm's computational complexity is called the number of arithmetic operations per iteration of the weight vector or coefficient vector. That is the number of multiplications, additions, and et.al. The multiplication operation's time-consuming operation is far greater than the addition operation's time-consuming operation, so the multiplication operation occupies a large proportion of the diffusion adaptive filtering algorithm's computational complexity. Therefore, computational complexity is an important property that affects the performance of the diffusion adaptive filtering algorithm. The DLLCLMS, DQQCLMS, and DLECLMS algorithms have two parameters: $a$ and $b$, during an exponential function in the DLECLMS algorithm, so the computational complexity of the DLLCLMS algorithm, the DQQCLMS algorithm, and the DLECLMS slightly larger than the DSELMS [23], DRVSSLMS [25], DLLAD [8] algorithms. Furthermore, when $M$ increases, those algorithms have the same computational complexity.

Table 2. The computational complexity of the DSELMS, DRVSSLMS, DLLAD, and three proposed diffusion adaptive filtering algorithms

| **Algorithm** | **Computational cost per iteration** | | | | | |
|---|---|---|---|---|---|---|
| | Recursion | × | + | \| \| | sign(·) | exp(·) |
| **DSELMS** | Eq. 1$a$ in [23] | $(2M+1)N+M$ | $(3M-1)N$ | 0 | $N$ | 0 |
| | Eq. 1$b$ in [23] | $NM$ | $(N-1)M$ | 0 | 0 | 0 |
| **DRVSSLMS** | Eq. 11 in [25] | $> (3M+1)N+M$ | $(3M-1)N$ | 0 | 0 | 0 |
| | Eq. 11 in [25] | $> (3M+1)N+M$ | $(3M-1)N$ | 0 | $N$ | 0 |
| | Eq. 12 in [25] | $NM$ | $(N-1)M$ | 0 | 0 | 0 |
| **DLLAD** | Eq. 16 in [8] | $2MN+M$ | $(3M-1)N$ | $N$ | 0 | 0 |
| | Eq. 17 in [8] | $NM$ | $(N-1)M$ | 0 | 0 | 0 |
| **DLLCLMS** | Eq. 16(I) in this paper | $(2M+2)N+M$ | $(3M-1)N$ | 0 | $N$ | 0 |
| | Eq. 16(II) in this paper | $(2M+2)N+M$ | $(3M-1)N$ | 0 | $N$ | 0 |
| | Eq. 17 in this paper | $NM$ | $(N-1)M$ | 0 | 0 | 0 |
| **DQQCLMS** | Eq. 18(I) in this paper | $(2M+3)N+M$ | $(3M-1)N$ | 0 | $N$ | 0 |
| | Eq. 18(II) in this paper | $(2M+3)N+M$ | $(3M-1)N$ | 0 | $N$ | 0 |
| | Eq. 19 in this paper | $NM$ | $(N-1)M$ | 0 | 0 | 0 |
| **DLECLMS** | Eq. 20 in this paper | $(2M+5)N+M$ | $3MN$ | 0 | 0 | $N$ |
| | Eq. 21 in this paper | $NM$ | $(N-1)M$ | 0 | 0 | 0 |

where: “×” denotes “Multiplications”; “>” denotes “larger than”; “+” denotes “Additions”; “| |” denotes “Absolute”.

### 3.3. Parameters $a$ and $b$ for the proposed algorithms

As described early, parameters $a$ and $b$ determine the shape and characteristics of each cost function because, as an essential core parameter, how to set it is very critical. So the choice of $a$ and $b$ in Eq. (16) ~ Eq. (21) plays a vital role in the DLLCLMS, DQQCLMS, and DLECLMS algorithms performance. The optimum cut-off value, $a$ and $b$, under different input signals, different impulsive noise, and different network structures, both the theory derivation method and simulation experimental method can be used. For the theory derivation method, although the optimal parameters $a$ and $b$ of the proposed three diffusion adaptive filtering algorithms are obtained based on minimizing the mean-square deviation (MSD) at the current time, the problem with this operation is that iterative formulas will increase the computational complexity. In this paper, the experimental simulation method will get the parameters $a$ and $b$ of the proposed three diffusion adaptive filtering algorithms. The following will explore the optimal parameters $a$ and $b$ of these three diffusion algorithms one by one. Simulation experiments with an unknown linear system, we set $M = 16$, and the parameters weight vector is selected randomly. Each distributed network topology consists of $N = 20$ nodes. For impulsive noises, in [35, 36], we can compute the impulsive noises by using the Levy alpha-stable distribution with setting $\alpha, \beta, \gamma, and\ \delta$ in MATLAB software (2016b). Besides, we set the impulsive noises is spatiotemporally independent. For the adaptation weights in the adaptation step and combination weights in the combination step, we apply the uniform rule i.e. $c_{l,n} = 1/N_n$. We use the network MSD to evaluate the performance of diffusion adaptive filtering algorithms, where $\mathrm{MSD}(i) = \frac{1}{N}\sum_{n=1}^{N} \mathrm{E}[|\mathbf{W}_o - \mathbf{W}_n(i)|^2]$. In addition, the independent Monte Carlo number is 20 and each run has 1000 iteration numbers.

### (1) For the DLLCLMS algorithm

**Parameter *a*.**

The choice of *a* in Eq. (16) plays a vital role in the DLLCLMS algorithm performance. Besides choosing the optimum cut-off value *a* under a variety of input signals, different intensities of impulsive noises, and different network structures, we set four groups of the experiment in a system identification application. We evaluate different $a$ estimators' relative efficiency based on their MSD using the DLLCLMS algorithm. Fig. 2 and Fig. 3, considering the convergence rate and the steady-state MSD, we know the DLLCLMS algorithm is robust for different probability densities of impulsive noises when *a*=0.8 and *b*=6.

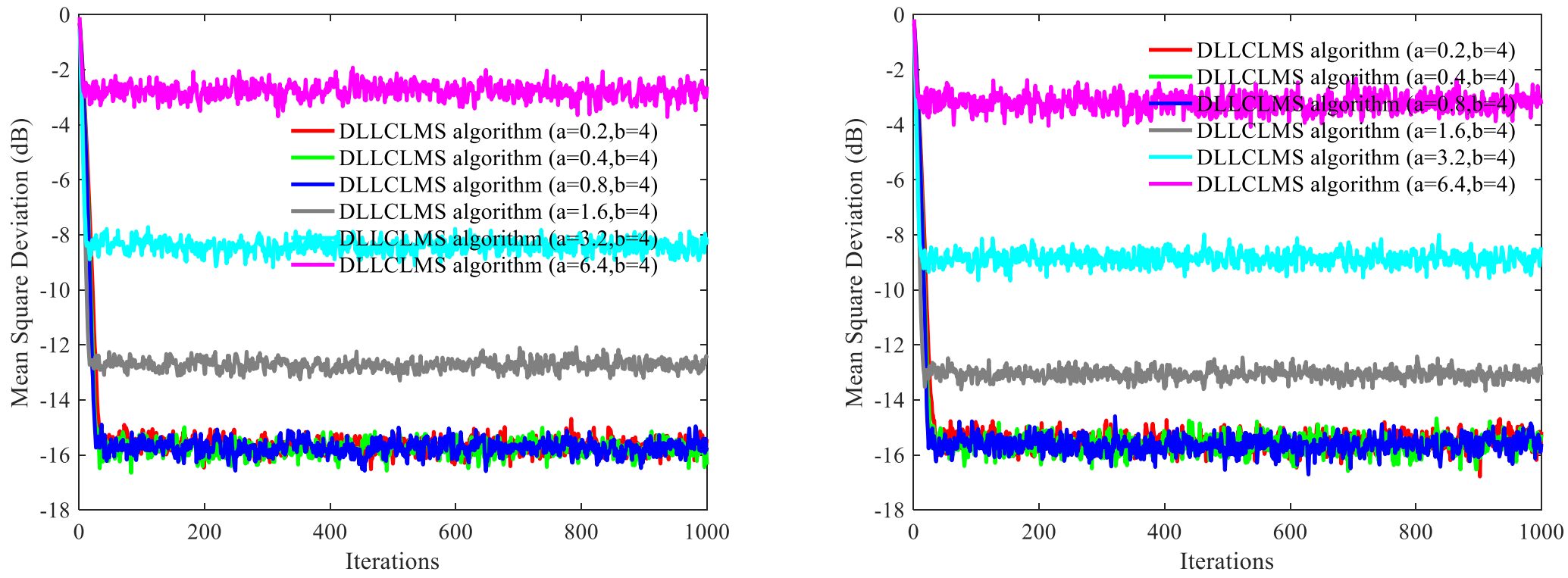

Fig.2 MSD curve with different *a* of the DLLCLMS algorithm ($\mu$=0.4) when network topology and neighbors to be decided by probability (probability=0.2) with $\alpha = 1.6, \beta = 0.05, \gamma = 0, and \ \delta = 1000$: (Left) $\mathbf{R}_{xx,n} = \sigma_{x,n}^2 \mathbf{I}_M$. (Right) $\mathbf{R}_{xx,n} = \sigma_{x,n}^2(i)\mathbf{I}_M, i = 1,2,3,\cdots, M$.

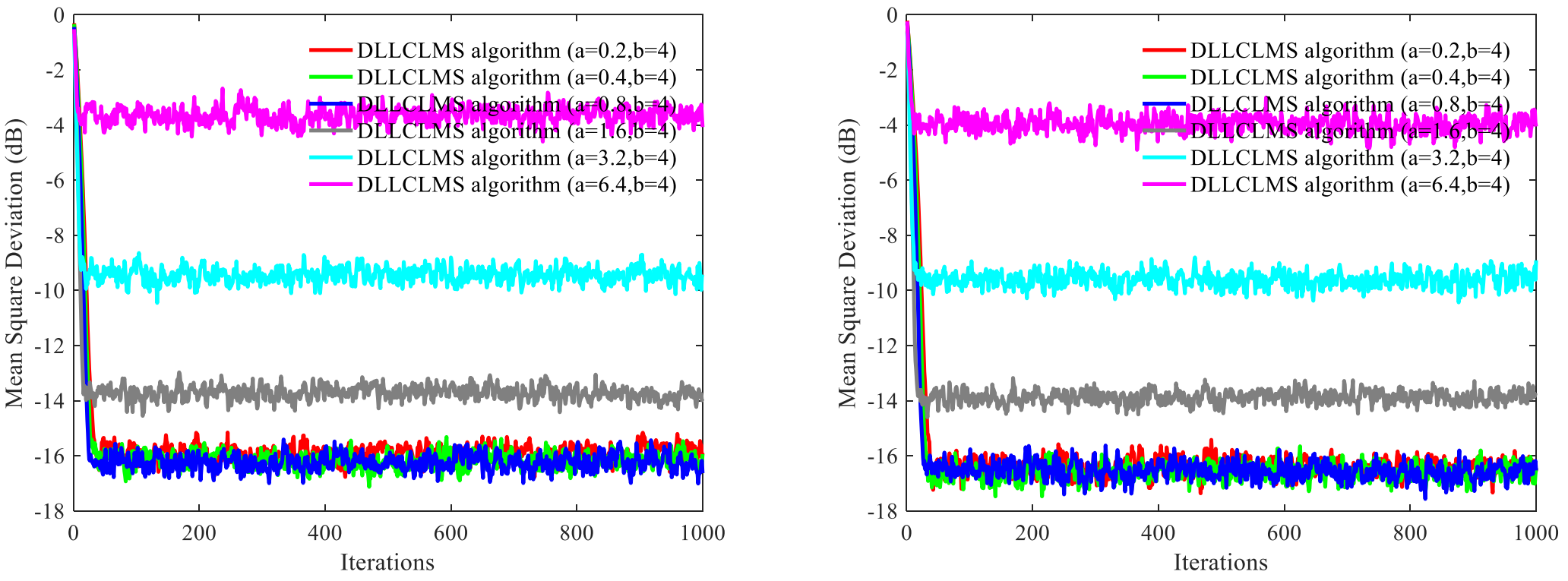

Fig.3 MSD curve with different *a* of the DLLCLMS algorithm ($\mu$=0.4) when network topology and neighbors to be decided by closeness in the distance (radius=0.3) with $\alpha = 1.2, \beta = 0.05, \gamma = 0, and \ \delta = 1000$: (Left) $\mathbf{R}_{xx,n}$ is a diagonal matrix with possibly different diagonal entries chosen randomly. (Right) $\mathbf{R}_{xx,n} = \sigma_{x,n}^2(i)\mathbf{I}_M, i = 1,2,3,\cdots, M$.

## Parameter *b*

The choice of *b* in Eq. (16) plays a vital role in the DLLCLMS algorithm performance. Besides choosing the optimum cut-off value *b* under different input signals, different intensities of impulsive noises, and different network structures, we set four groups of the experiment in a system identification application. We evaluate the relative efficiency of different *b* estimators' relative efficiency based on their MSD using the DLLCLMS algorithm. Fig. 4 and Fig. 5, considering the convergence rate and the steady-state MSD, we know the DLLCLMS algorithm is robust for different probability densities of impulsive noises when *b*=4 and *a*=0.8.

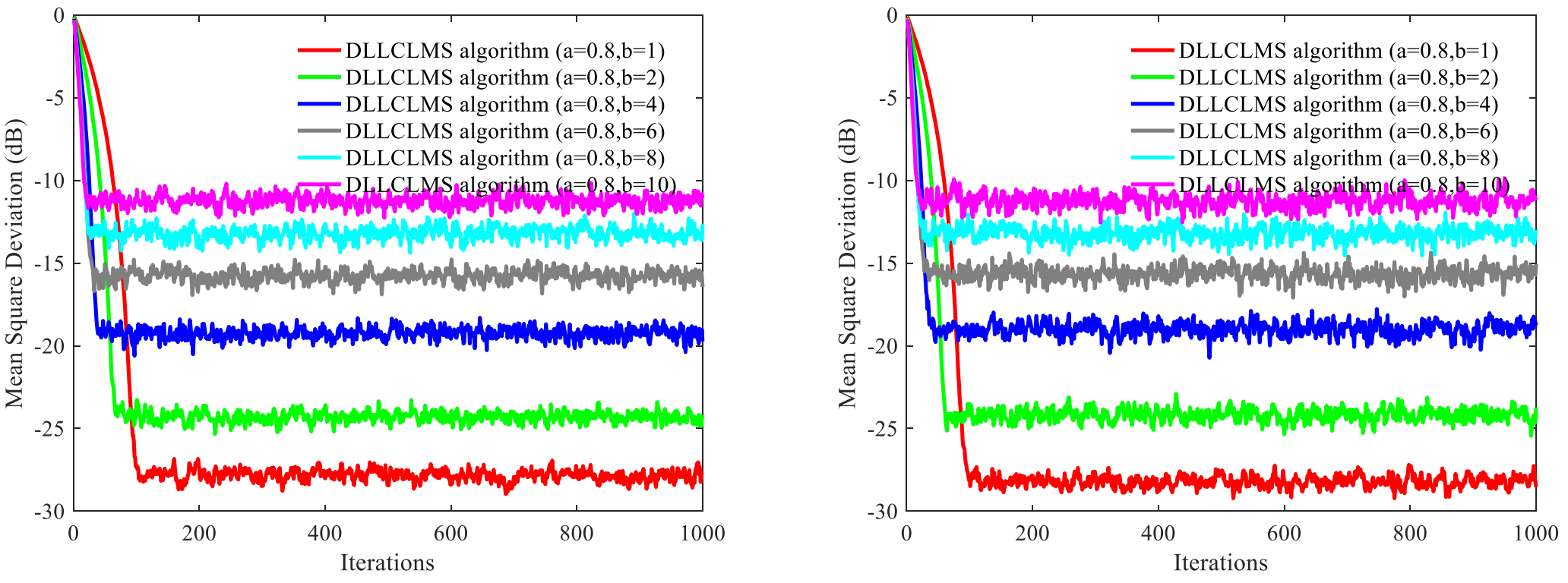

Fig.4 MSD curve with different *b* of the DLLCLMS algorithm ($\mu$=0.4) when network topology and neighbors to be decided by probability (probability=0.2) with $\alpha = 1.6, \beta = 0.05, \gamma = 0, and \ \delta = 1000$: (Left) $\mathbf{R}_{xx,n} = \sigma_{x,n}^2 \mathbf{I}_M$ and Pr=0.8. (Right) $\mathbf{R}_{xx,n} = \sigma_{x,n}^2(i)\mathbf{I}_M, i = 1,2,3,\cdots, M$.

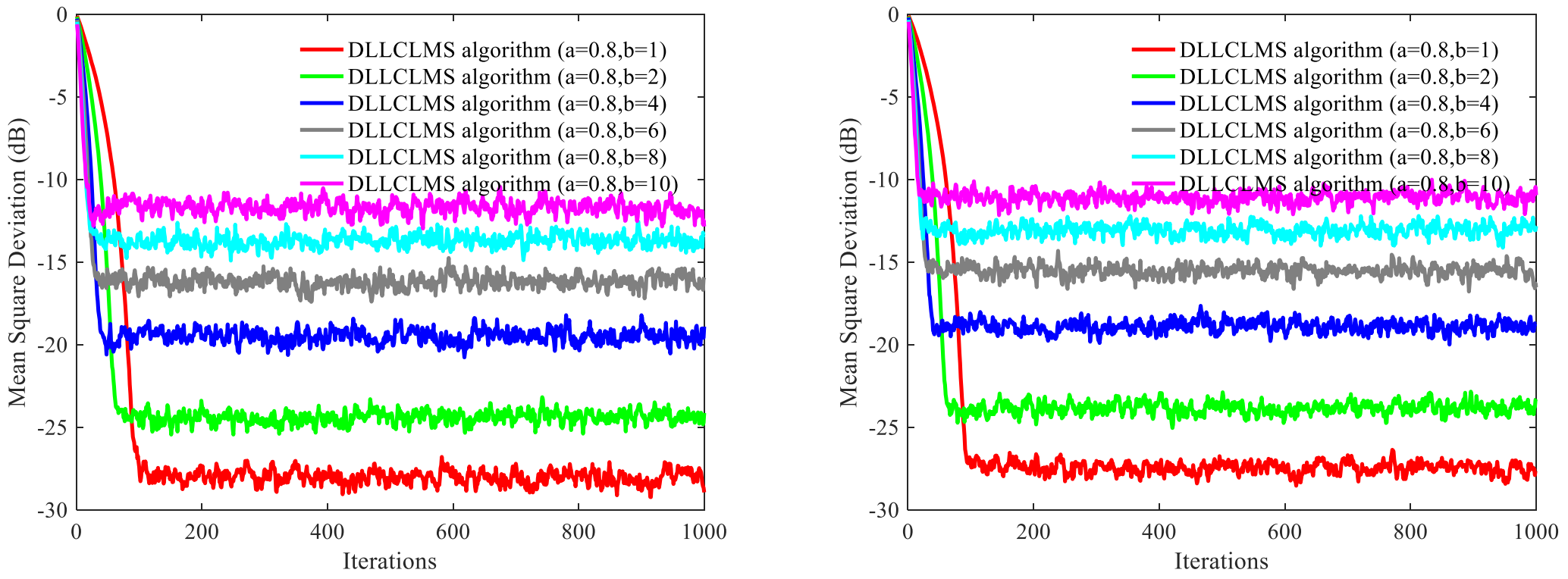

Fig.5 MSD curve with different $b$ of the DLLCLMS algorithm ($\mu$=0.4) when network topology and neighbors to be decided by closeness in distance (radius=0.3) with $\alpha = 1.2, \beta = 0.05, \gamma = 0, and\ \ \delta = 1000$: (Left) $\mathbf{R}_{xx,n}$ is a diagonal matrix with possibly different diagonal entries chosen randomly. (Right) $\mathbf{R}_{xx,n} = \sigma_{x,n}^2(i)\mathbf{I}_M, i = 1,2,3,\cdots, M$.

## (2) For the DQQCLMS algorithm

### Parameter *a*.

The choice of $a$ in Eq. (18) plays a vital role in the DQQCLMS algorithm's performance. Besides choosing the optimum cut-off value $a$ under different input signals, different intensities of impulsive noises, and different network structures, we set four groups of the experiment in a system identification application. We evaluate different $a$ estimators' relative efficiency based on their MSD using the DQQCLMS algorithm. Fig. 6 and Fig. 7, considering the convergence rate and the steady-state MSD, we know the DQQCLMS algorithm is robust for different probability densities of impulsive noises when $a$=0.8 with $b$=6.

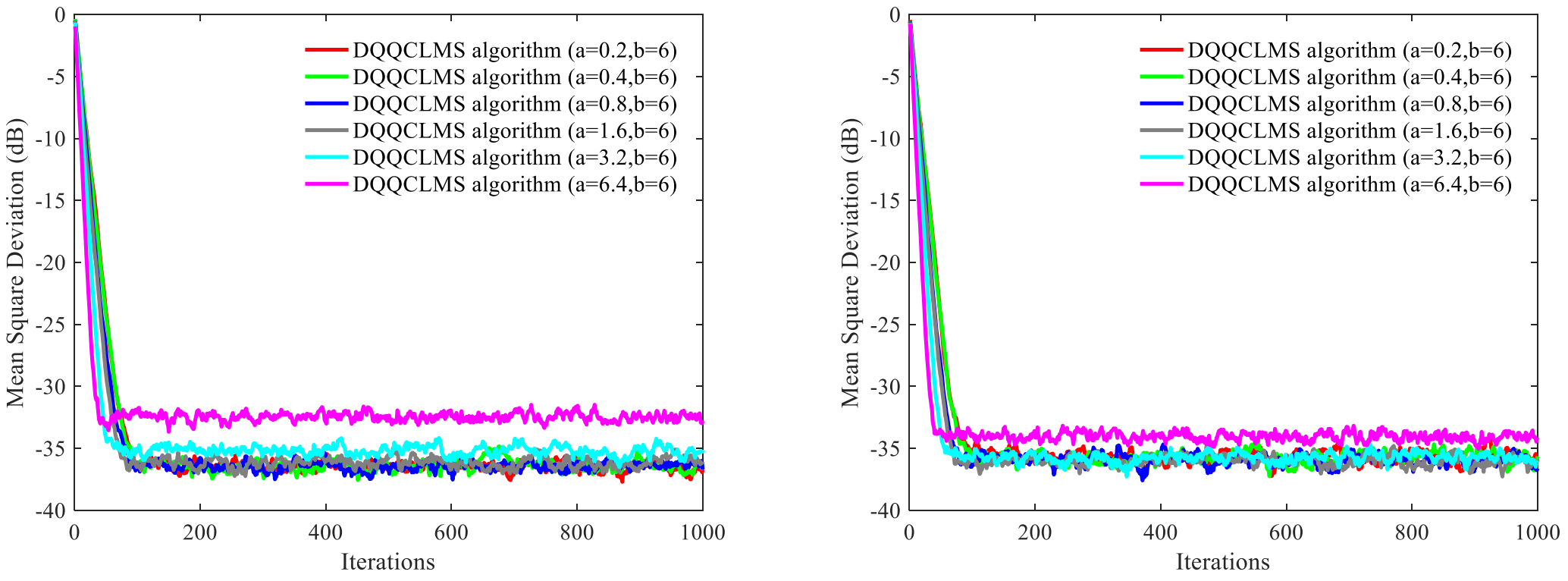

Fig.6 MSD curve with different $a$ of the DQQCLMS algorithm ($\mu$=0.4) when network topology and neighbors to be decided by probability (probability=0.2) with $\alpha = 1.6, \beta = 0.05, \gamma = 0, and\ \ \delta = 1000$: (Left) $\mathbf{R}_{xx,n} = \sigma_{x,n}^2\mathbf{I}_M$. (Right) $\mathbf{R}_{xx,n} = \sigma_{x,n}^2(i)\mathbf{I}_M, i = 1,2,3,\cdots, M$.

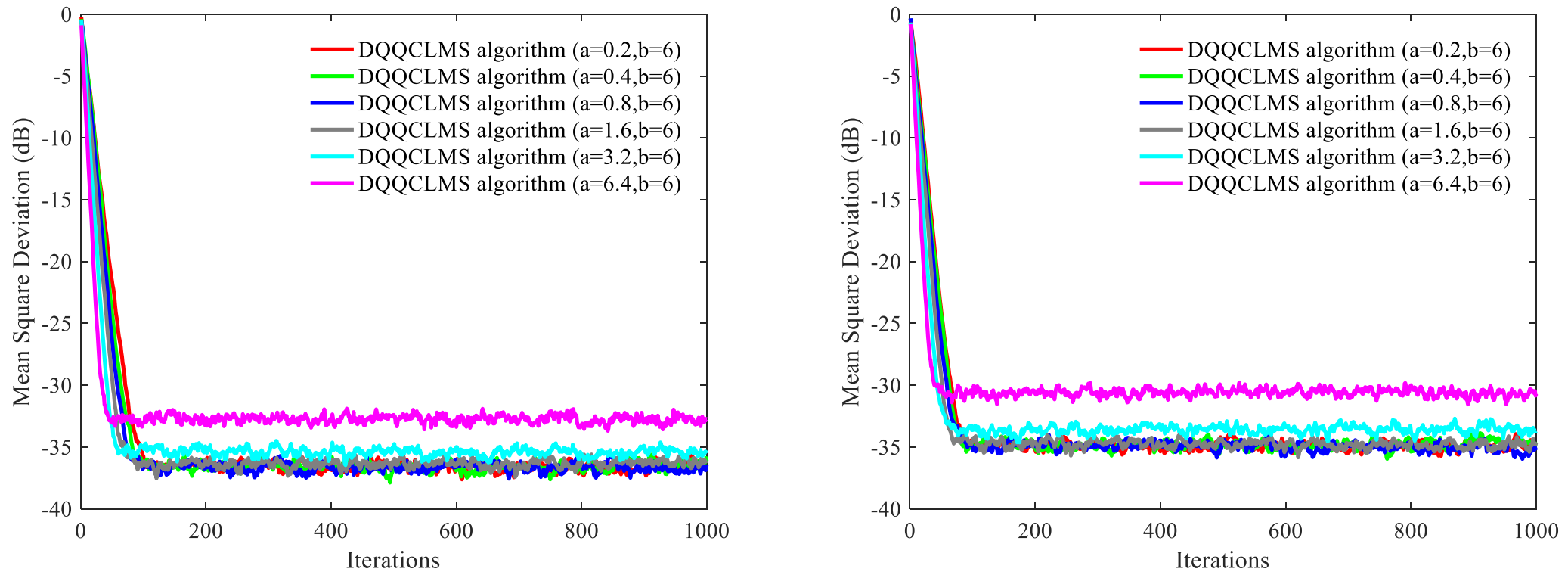

Fig.7 MSD curve with different *a* of the DQQCLMS algorithm ($\mu$=0.4) when network topology and neighbors to be decided by closeness in distance (radius=0.3) with $\alpha = 1.2, \beta = 0.05, \gamma = 0, and \;\; \delta = 1000$: (Left) $\mathbf{R}_{xx,n}$ is a diagonal matrix with possibly different diagonal entries chosen randomly. (Right) $\mathbf{R}_{xx,n} = \sigma_{x,n}^2(i)\mathbf{I}_M, i = 1,2,3,\cdots, M$.

## Parameter *b*

The choice of *b* in Eq. (18) plays a vital role in the DQQCLMS algorithm performance. Besides choosing the optimum cut-off value *b* under different input signals, different intensities of impulsive noises, and different network structures, we set four groups of the experiment in a system identification application. We evaluate the relative efficiency of different *b* estimators based on their MSD using the DQQCLMS algorithm. Fig. 8 and Fig. 9, considering the convergence rate and the steady-state MSD, we know the DQQCLMS algorithm is robust for different probability densities of impulsive noises when *b*=6 and *a*=0.8.

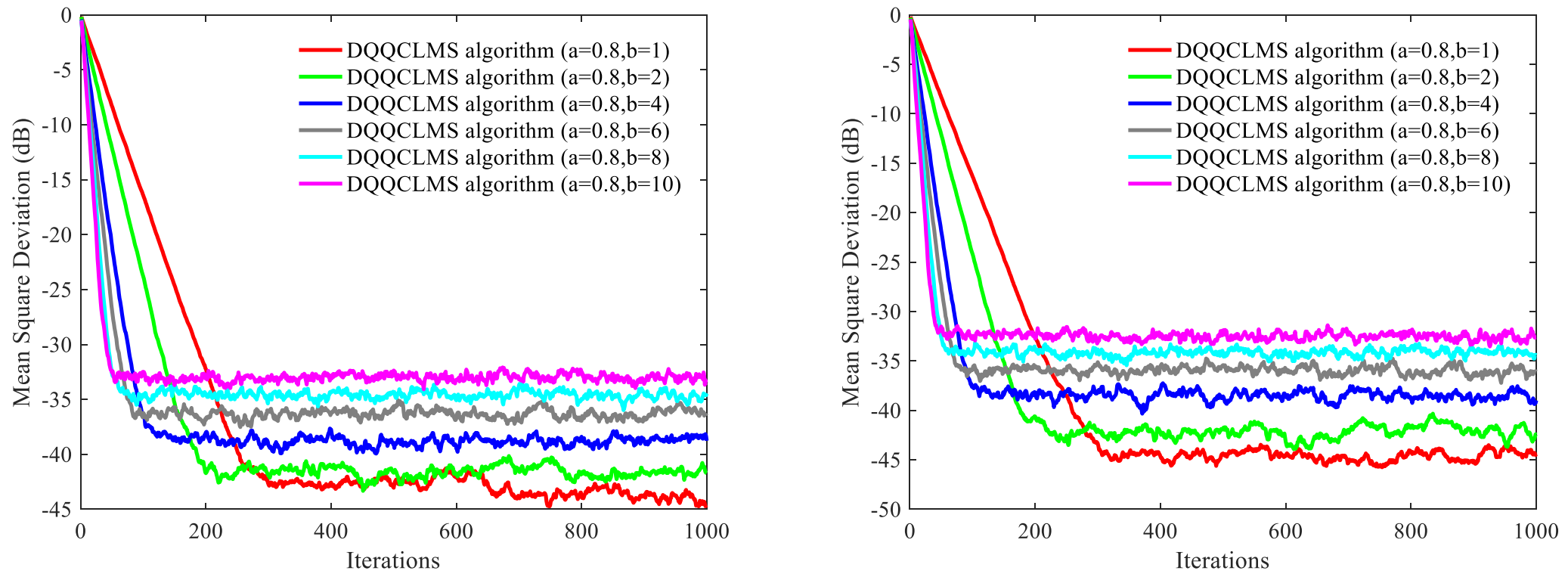

Fig.8 MSD curve with different *b* of the DQQCLMS algorithm ($\mu$=0.4) when network topology and neighbors to be decided by probability (probability=0.2) with $\alpha = 1.6, \beta = 0.05, \gamma = 0, and \;\; \delta = 1000$: (Left) $\mathbf{R}_{xx,n} = \sigma_{x,n}^2\mathbf{I}_M$ and Pr=0.8. (Right) $\mathbf{R}_{xx,n} = \sigma_{x,n}^2(i)\mathbf{I}_M, i = 1,2,3,\cdots, M$.

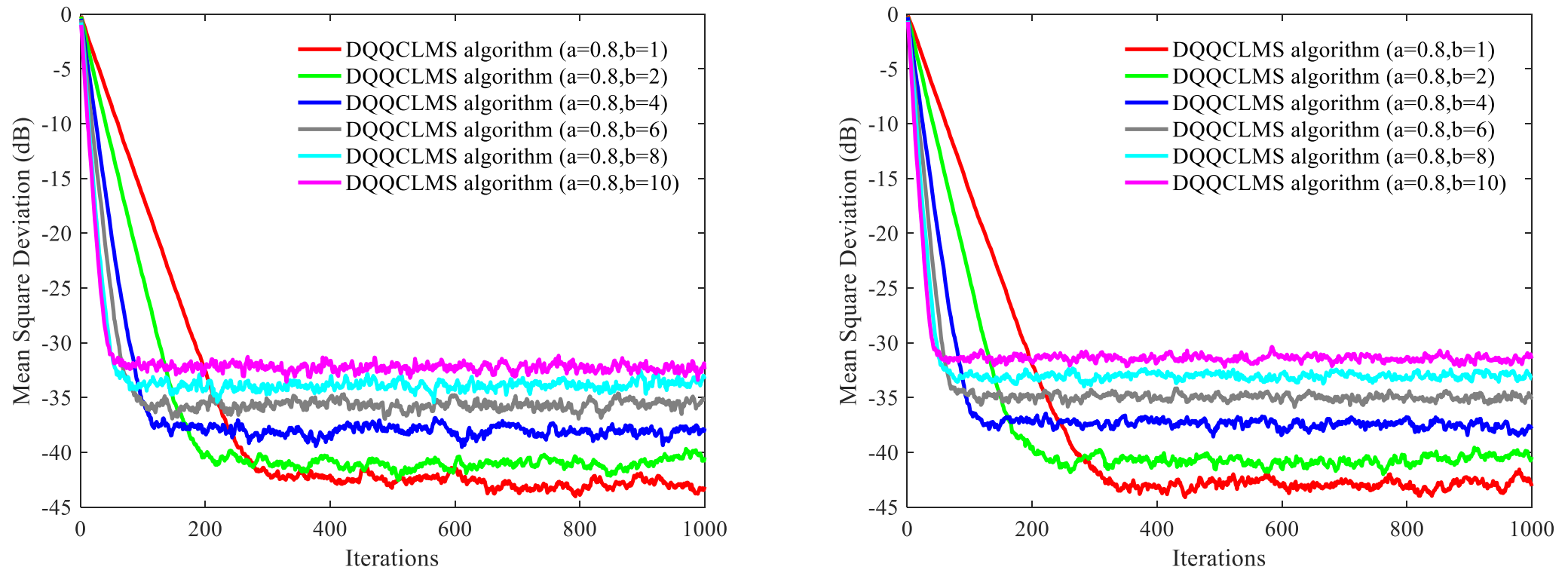

Fig.9 MSD curve with different $b$ of the DQQCLMS algorithm ($\mu$=0.4) when network topology and neighbors to be decided by closeness in distance (radius=0.3) with $\alpha = 1.2, \beta = 0.05, \gamma = 0, and\ \ \delta = 1000$: (Left) $\mathbf{R}_{xx,n}$ is a diagonal matrix with possibly different diagonal entries chosen randomly. (Right) $\mathbf{R}_{xx,n} = \sigma_{x,n}^2(i)\mathbf{I}_M, i = 1,2,3,\cdots,M$.

## (3) For the DLECLMS algorithm

**Parameter *a*.**

The choice of $a$ in Eq. (20) plays a vital role in the DLECLMS algorithm performance. Besides choosing the optimum cut-off value $a$ under different input signals, different intensities of impulsive noises, and different network structures, we set four groups of the experiment in a system identification application. We evaluate different $a$ estimators' relative efficiency based on their MSD using the DLECLMS algorithm. Fig. 10 and Fig. 11, considering the convergence rate and the steady-state MSD, we know the DLECLMS algorithm is robust for different probability densities of impulsive noises when $a$=0.32 and $b$=6.

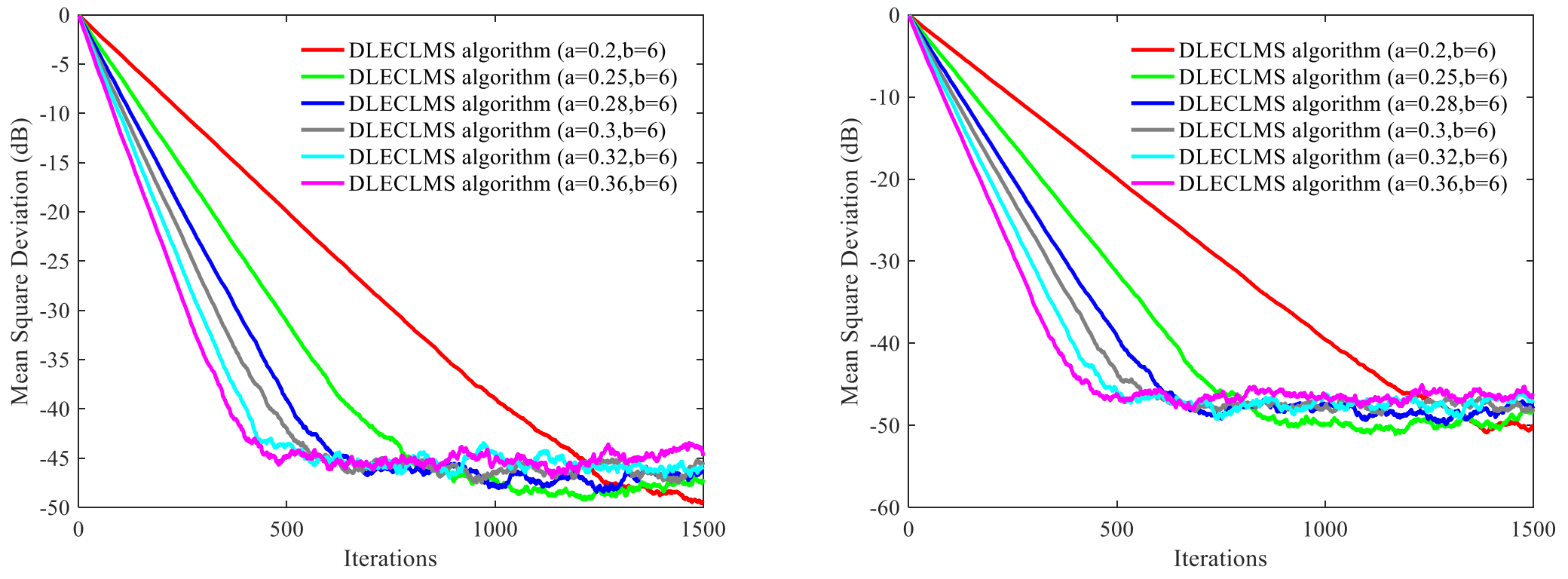

Fig.10 MSD curve with different $a$ of the DLECLMS algorithm ($\mu$=0.4) when network topology and neighbors to be decided by probability (probability=0.2) with $\alpha = 1.6, \beta = 0.05, \gamma = 0, and\ \ \delta = 1500$: (Left) $\mathbf{R}_{xx,n} = \sigma_{x,n}^2\mathbf{I}_M$. (Right) $\mathbf{R}_{xx,n} = \sigma_{x,n}^2(i)\mathbf{I}_M, i = 1,2,3,\cdots,M$.

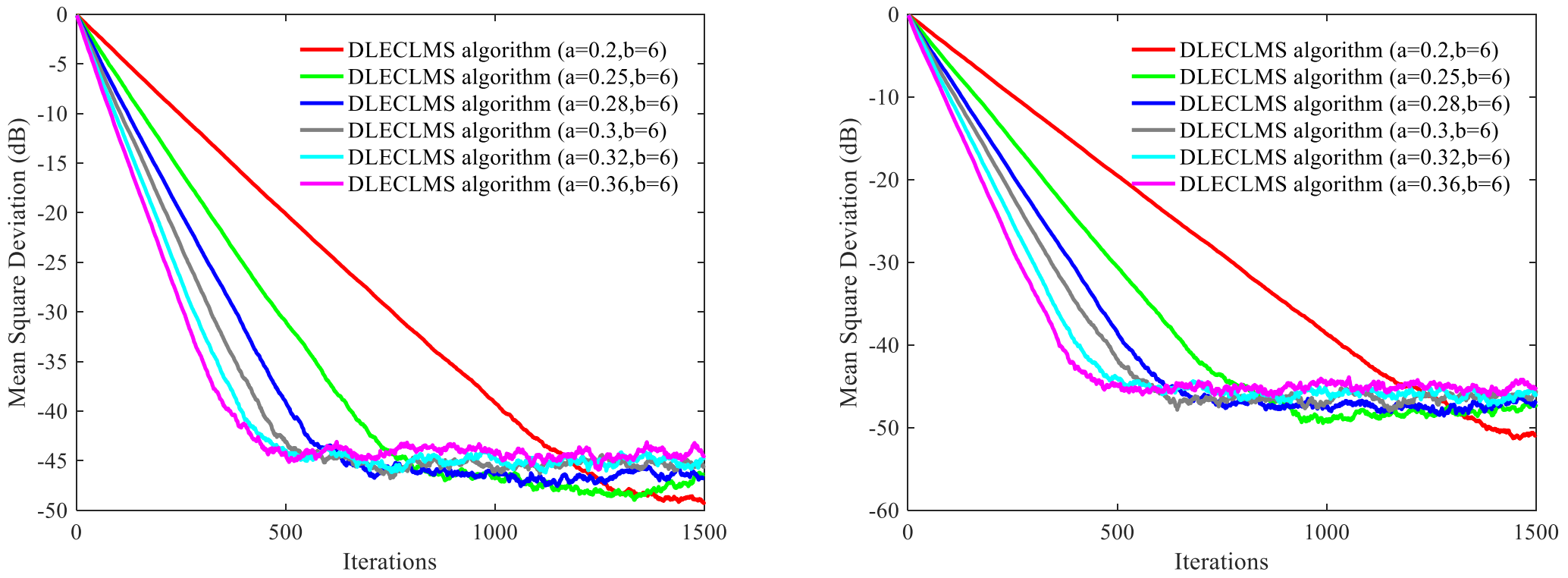

Fig.11 MSD curve with different $a$ of the DLECLMS algorithm ($\mu$=0.4) when network topology and neighbors to be decided by closeness in distance (radius=0.3) with $\alpha = 1.2, \beta = 0.05, \gamma = 0, and \ \ \delta = 1500$: (Left) $\mathbf{R}_{xx,n}$ is a diagonal matrix with possibly different diagonal entries chosen randomly. (Right) $\mathbf{R}_{xx,n} = \sigma_{x,n}^2(i)\mathbf{I}_M, i = 1,2,3,\cdots,M$.

## Parameter *b*

The choice of $b$ in Eq. (20) plays a vital role in the DLECLMS algorithm performance. Besides choosing the optimum cut-off value $b$ under different input signals, different intensities of impulsive noises, and different network structures, we set four groups of the experiment in a system identification application. We evaluate the relative efficiency of different $b$ estimators based on their MSD using the DLECLMS algorithm. Fig. 12 and Fig. 13, considering the convergence rate and the steady-state MSD, we know the DLECLMS algorithm is robust for different probability densities of impulsive noises when $b$=6 and $a$=0.32.

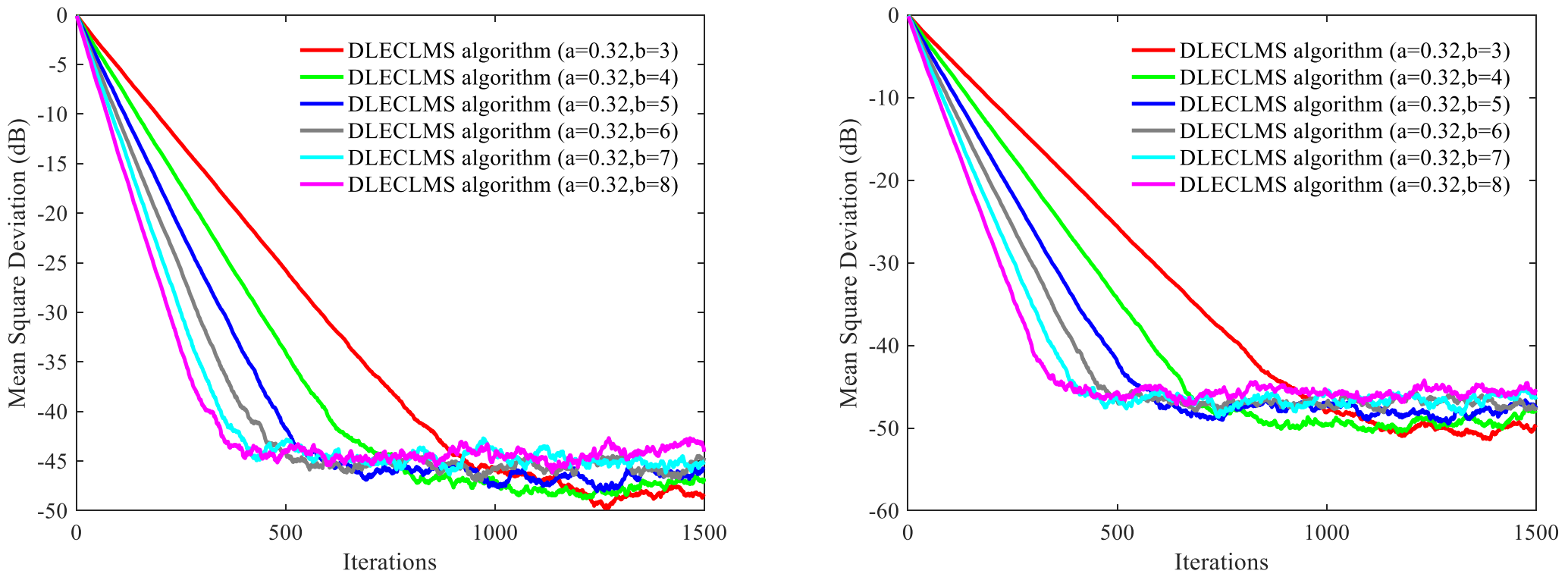

Fig.12 MSD curve with different $b$ of the DLECLMS algorithm ($\mu$=0.4) when network topology and neighbors to be decided by probability (probability=0.2) with $\alpha = 1.6, \beta = 0.05, \gamma = 0, and \ \ \delta = 1500$ : (Left) $\mathbf{R}_{xx,n} = \sigma_{x,n}^2\mathbf{I}_M$ and Pr=0.8. (Right) $\mathbf{R}_{xx,n} = \sigma_{x,n}^2(i)\mathbf{I}_M, i = 1,2,3,\cdots,M$.

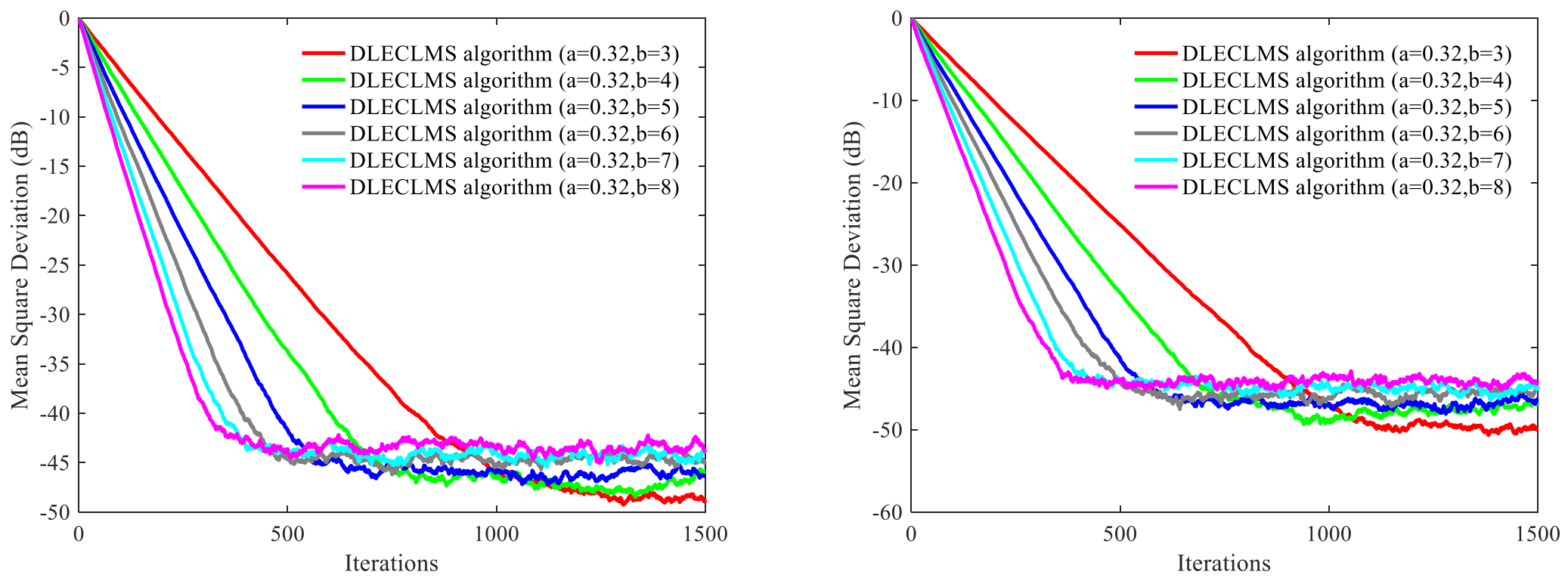

Fig.13 MSD curve with different $b$ of the DLECLMS algorithm ($\mu$=0.4) when network topology and neighbors to be decided by closeness in distance (radius=0.3) with $\alpha = 1.2, \beta = 0.05, \gamma = 0, and\ \delta = 1500$: (Left) $\mathbf{R}_{xx,n}$ is a diagonal matrix with possibly different diagonal entries chosen randomly. (Right) $\mathbf{R}_{xx,n} = \sigma_{x,n}^2(i)\mathbf{I}_M, i = 1,2,3,\cdots, M$.

## 4. Simulation results

Because in this paper, we focus on the distributed adaptive filter algorithm and compare the DLLCLMS, DQQCLMS, and DLECLMS algorithms with the DSELMS [23], DRVSSLMS [25], and DLLAD [8] algorithms in linear system identification. Then, we want to demonstrate the robustness performance of the three proposed DLLCLMS, DQQCLMS, and DLECLMS algorithms in the presence of different intensities of impulsive noise and different input signals. Several group simulation experiments with different intensities of impulsive noises and different input signal types are set. For an unknown linear system, we set $M = 16$, and the parameters weight vector is selected randomly. Each distributed network topology consists of $N = 20$ nodes. For impulsive noises, in [35, 36], we can compute the impulsive noises by using the Levy alpha-stable distribution with setting $\alpha, \beta, \gamma, and\ \delta$. Besides, we set the impulsive noises is spatiotemporally independent. For the adaptation weights in the adaptation step and combination weights in combination step, we apply the uniform rule i.e. $c_{l,n} = 1/N_n$. We use the network MSD to evaluate the performance of diffusion adaptive filtering algorithms, where $\text{MSD}(i) = \frac{1}{N}\sum_{n=1}^{N} \text{E}[|\mathbf{W}_o - \mathbf{W}_n(i)|^2]$. In addition, the independent Monte Carlo number is 20 and each run has 2000 iteration numbers.

***Simulation experiment 1***

The convergence rate is faster, and the steady-state MSD is lower to show the distributed adaptive filter algorithms we proposed are more robust to the input signal than DSELMS, DRVSSLMS, and DLLAD. Set up three experiments; both have the same network topology, the same impulsive noise, and different input signals. If any two network topology nodes are declared neighbors, the connection probability is greater than or equal to 0.2, and the network topology is shown in Fig.14. The MSD iteration curves for DRVSSLMS ($\mu$ equal to 0.35), DSELMS ($\mu$ equal to 0.35), DNLMS ($\mu$ equal to 0.35), and DLLAD ($\mu$ equal to 0.35) algorithms in Fig.15, Fig.16, and Fig.17 are different types of the input signal when the measurement noise in an unknown system is impulsive noises with $\alpha = 1.6, \beta = 0.05, \gamma = 0, and \ \ \delta = 2000$.

Fig. 15, Fig.16, and Fig.17 show that although different input signals are used, the DLLCLMS, DQQCLMS, and DLECLMS algorithms have a faster convergence rate and lower steady-state MSD than the DSELMS, DRVSSLMS, and DLLAD algorithms. Besides, the DLLCLMS, DQQCLMS, and DLECLMS algorithms are more robust to the input signal. In conclusion, from ***Simulation experiment 1***, we can get the DLLCLMS, DQQCLMS, and DLECLMS algorithms are superior to the DSELMS, DRVSSLMS, and DLLAD algorithms. Furthermore, the order of performance superiority is DLECLMS, DQQCLMS, and DLLCLMS.

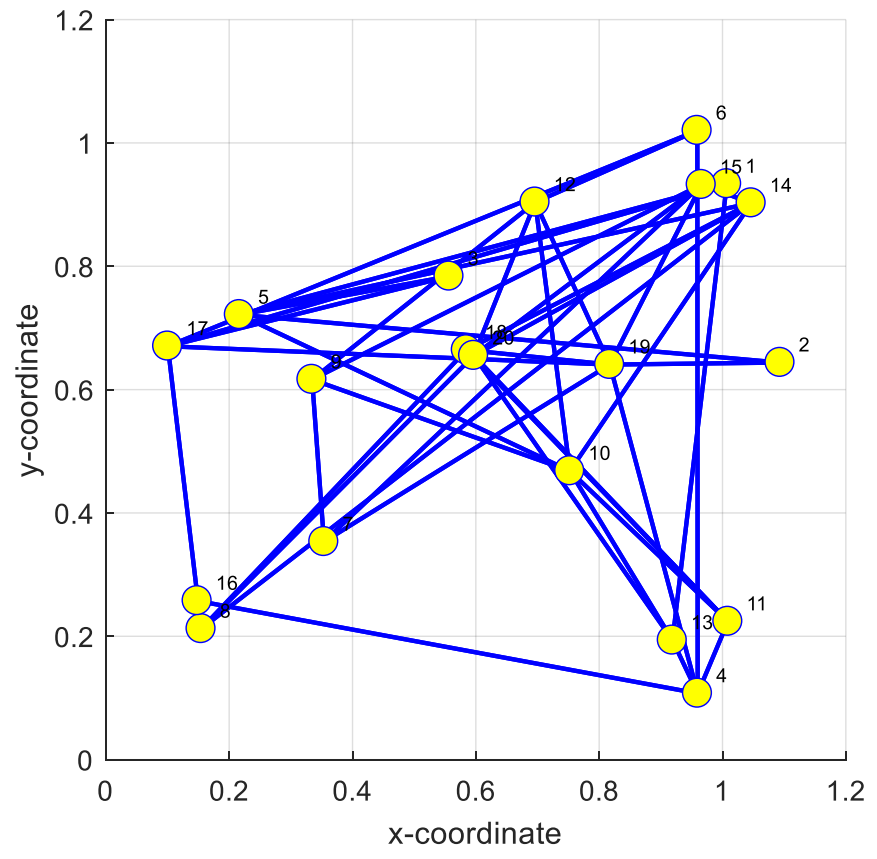

Fig.14. Random network topology to be decided by probability.

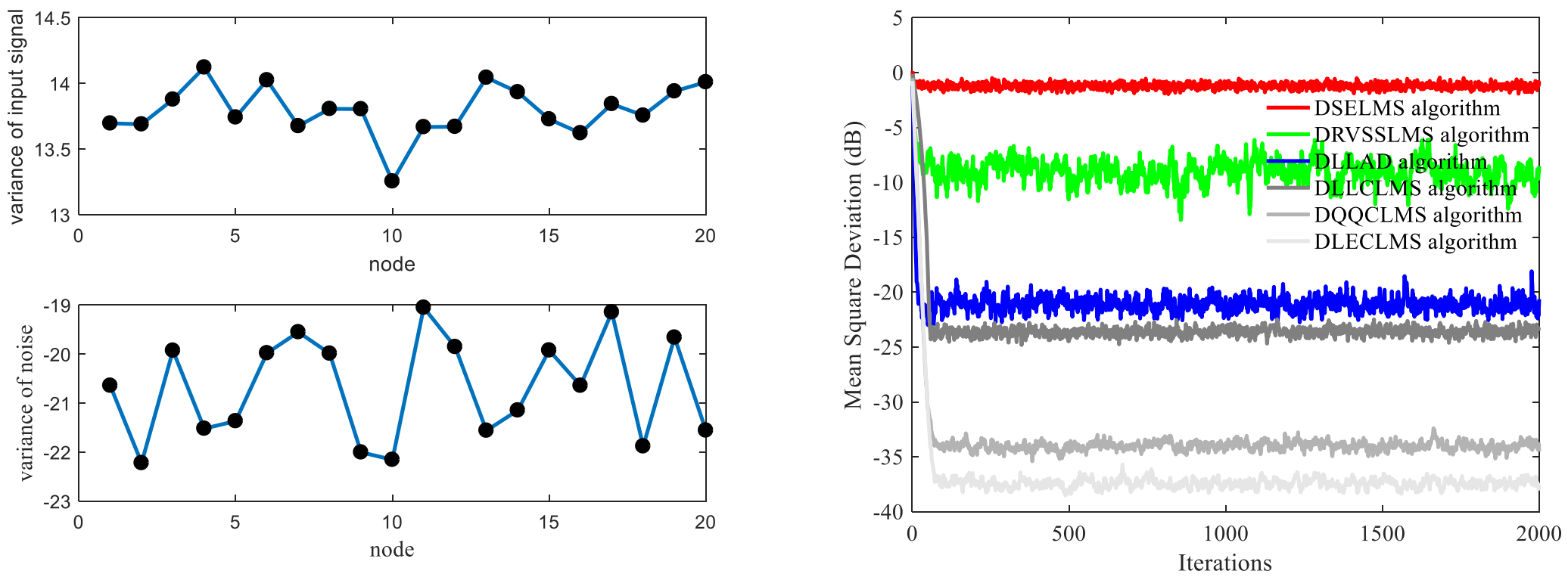

Fig.15. (Left_top) the input signals $\{\mathbf{X}_n(i)\}$ variances of at each network node with $\mathbf{R}_{xx,n} = \sigma_{x,n}^2 \mathbf{I}_M$ with possibly different diagonal entries chosen randomly, (Left_bottom) the measurement noise variances $\varepsilon_n(i)$ at each network node; (Right) Transient network MSD (dB) iteration curves of the DSELMS, DRVSSLMS, DLLAD, DLLCLMS, DQQCLMS, and DLECLMS algorithms.

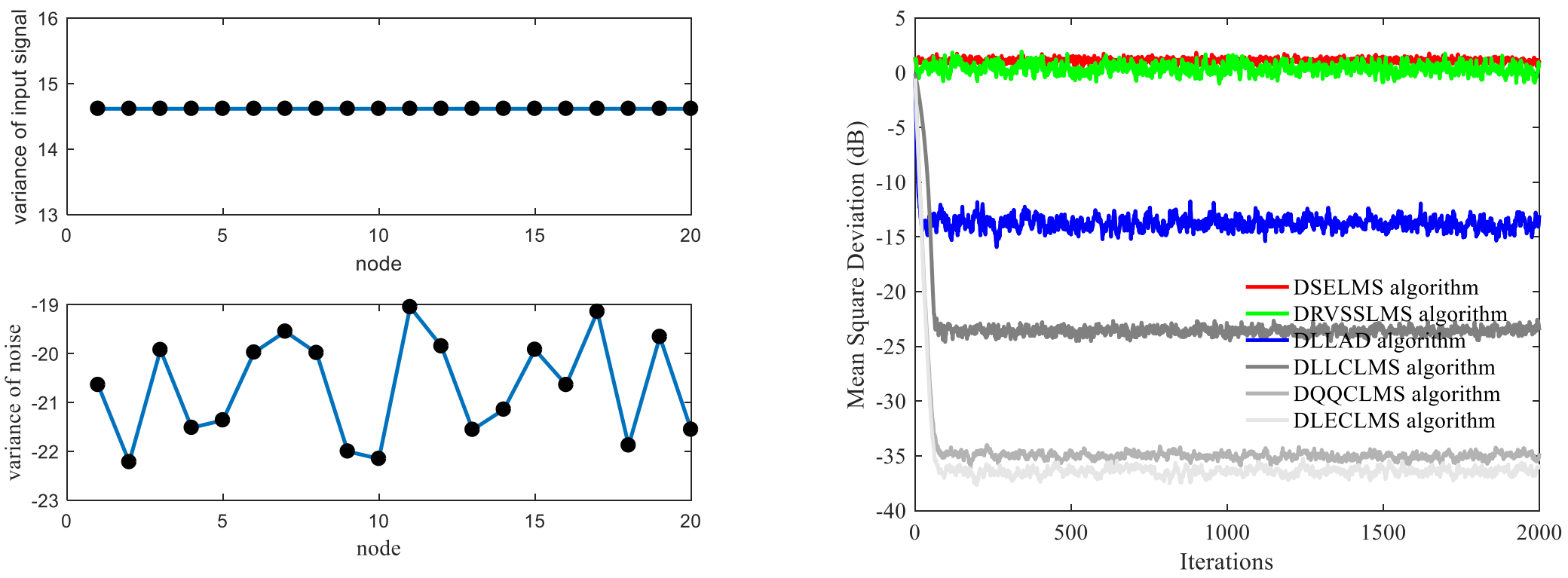

Fig.16. (Left_top) the input signals $\{\mathbf{X}_n(i)\}$ variances of at each network node with $\mathbf{R}_{xx,n} = \sigma_{x,n}^2 \mathbf{I}_M$ with the same value in each diagonal entries, (Left_bottom) the measurement noise variances $\{\varepsilon_n(i)\}$ at each network node; (Right) Transient network MSD (dB) iteration curves of the DSELMS, DRVSSLMS, DLLAD, DLLCLMS, DQQCLMS, and DLECLMS algorithms.

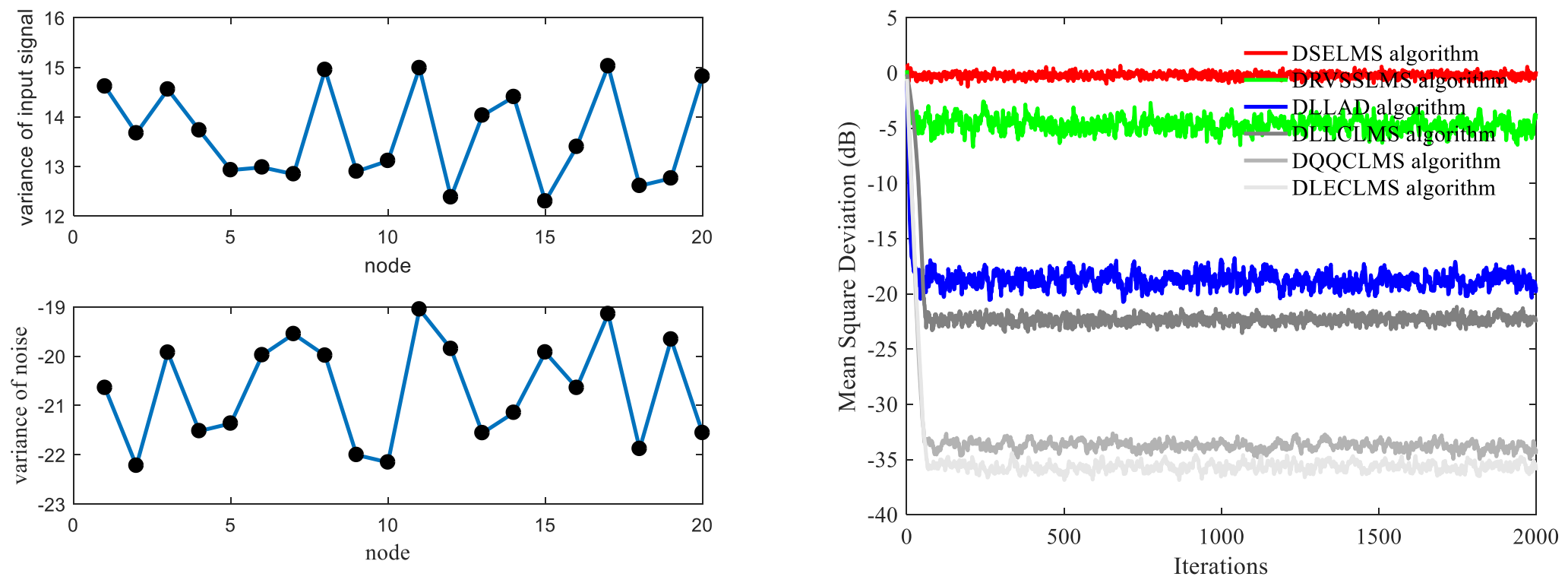

Fig.17. (Left_top) the input signals $\{\mathbf{X}_n(i)\}$ variances of at each network node with $\mathbf{R}_{xx,n} = \sigma_{x,n}^2(t)\mathbf{I}_M, t = 1,2,3,\cdots,M$ with a different value in each diagonal entries, (Left_bottom) the measurement noise variances $\{\varepsilon_n(i)\}$ at each network node; (Right) Transient network MSD (dB) iteration curves of the DSELMS, DRVSSLMS, DLLAD, DLLCLMS, DQQCLMS, and DLECLMS algorithms.

### *Simulation experiment 2*

The convergence rate is faster, and the steady-state MSD is lower to show the distributed adaptive filter algorithms we proposed are more robust to the impulsive noise than DSELMS, DRVSSLMS, and DLLAD. Set up three experiments; both have the same network topology, the same input signal, and different intensities of impulsive noises. If any two network topology nodes are declared neighbors, a certain radius for each node is larger than or equal to 0.3; the network topology is shown in Fig. 18(Left). The MSD iteration curves for DRVSSLMS ($\mu$ equal to 0.35), DSELMS ($\mu$ equal to 0.35), DNLMS ($\mu$ equal to 0.35), DLLAD ($\mu$ equal to 0.35), DLLCLMS, DQQCLMS, and DLECLMS algorithms in Fig.19 with $\alpha = 1.6$, $\alpha = 1.1$, $\alpha = 0.8$, and $\alpha = 0.4$ with $\beta = 0.05, \gamma = 0, and\ \ \delta = 2000$, respectively.

From Fig. 19, we can find that although different probability density of impulsive noise is considered, the DLLCLMS, DQQCLMS, and DLECLMS algorithms have a slightly faster rate than that of the DSELMS, DRVSSLMS, and DLLAD algorithms.

The DLLCLMS, DQQCLMS, and DLECLMS algorithms still have a minor steady-state error than the DSELMS, DRVSSLMS, and DLLAD algorithms. In a word, from ***Simulation experiment 2***, we can observe that the DLLCLMS, DQQCLMS, and DLECLMS algorithms are more robust to impulsive noise than the DSELMS, DRVSSLMS, and DLLAD algorithms. Furthermore, the order of performance superiority is DLECLMS, DQQCLMS, and DLLCLMS. And when noise distribution tends to Gaussian distribution, DLECLMS and DQQCLMS tend to be the same but better than DLLCLMS.

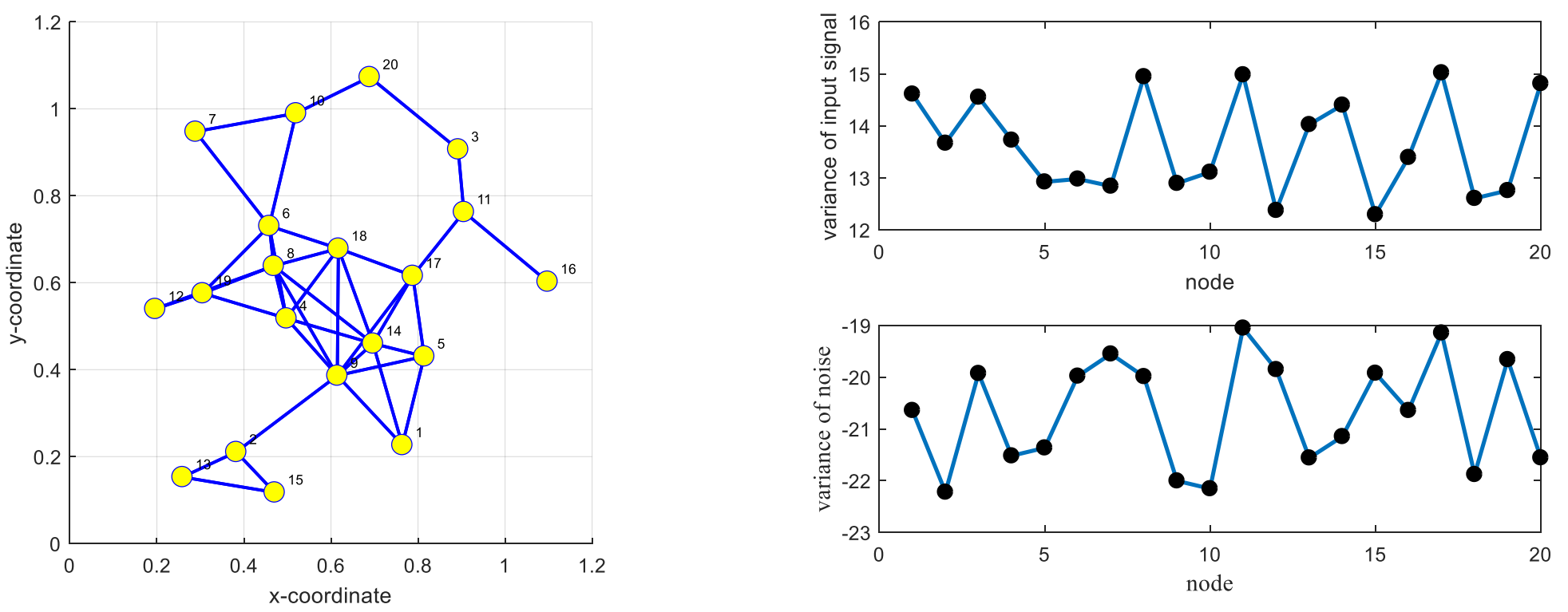

Fig.18. (Left) Random network topology to be decided by a certain radius; (Right_top) the input signals $\{\mathbf{X}_n(i)\}$ variances of at each network node with $\mathbf{R}_{xx,n} = \sigma_{x,n}^2 \mathbf{I}_M$ with possibly different diagonal entries chosen randomly, (Right_bottom) the measurement noise variances $\varepsilon_n(t)$ at each network node.

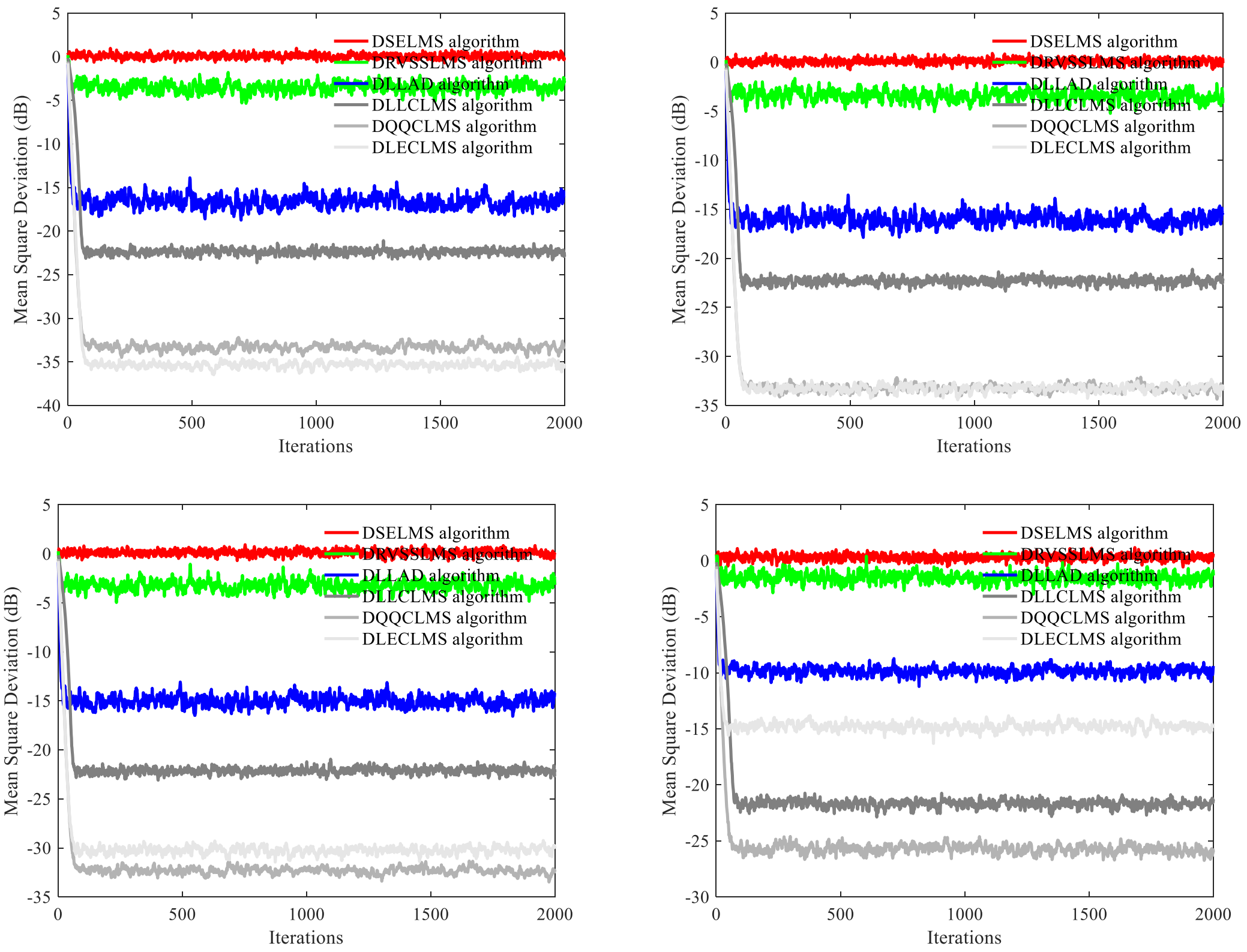

Fig.19. Transient network MSD (dB) iteration curves of the DSELMS, DRVSSLMS, DLLAD, DLLCLMS, DQQCLMS, and DLECLMS algorithms. (Up_Left) with $\alpha = 1.6$, (Up_right) with $\alpha = 1.1$, and (Down_left) with $\alpha = 0.8$, and (Down_Right) with $\alpha = 0.4$.

## 5. Conclusion

This paper proposed a family of diffusion adaptive filtering algorithms using three asymmetric costs of error functions; those three distributed adaptive filtering algorithms robust to the input signal and impulsive noise. Specifically, those three distributed adaptive algorithms are developed by modifying the DLMS algorithm and combining the LLC, QQC, and LEC functions at all distributed network nodes, then the DLLCLMS, DQQCLMS, and DLECLMS algorithms are proposed. The theoretical analysis demonstrates that those three distributed adaptive filtering algorithms can effectively estimate from an asymmetric cost function perspective. Besides, theoretical mean behavior interpreted that those three algorithms can achieve accurate estimation. Simulation results showed that the DLLCLMS, DQQCLMS, and DLECLMS algorithms are more robust to the input signal and impulsive noise than the DSELMS, DRVSSLMS, and DLLAD algorithms. That is to say, the DLLCLMS, DQQCLMS, and DLECLMS algorithms have superior performance when estimating the unknown linear system under the changeable impulsive noise environments and different types of input signals, which will have a significant impact on real-world applications. Besides, the environment in actual applications is more complex, nonlinear, and time-varying and needs to be adjusted accordingly for different application scenarios [6]. We also need to consider whether the system parameters to be evaluated are sparsity (such as brain network based on fMRI or EEG signals) [37]. In such cases, it will be better to add a regularized constraint terms to those adaptive algorithms.

## Acknowledgments

The author would like to thank Biswal Bharat and Chun Meng (School of Life Science and Technology, University of Electronic Science and Technology of China, Chengdu, China.) and Dr. Yang Qiu (Key Laboratory of State Ethnic Affairs Commission for Electronic and Information Engineering, Southwest Minzu University, Chengdu, China.) for thoughtful comments on a draft manuscript.